\pdfoutput=1

\documentclass[11pt]{article}

\usepackage[]{ACL2023}

\usepackage{times}
\usepackage{latexsym}
\usepackage{bbding}
\usepackage{graphicx}
\usepackage{subfigure}
\usepackage{caption}
\usepackage{amsmath}
\usepackage{tablefootnote}
\usepackage{threeparttable}
\usepackage{booktabs}
\usepackage{tabularx}
\usepackage{adjustbox}
\usepackage{algorithm}
\usepackage{algorithmic}
\usepackage{multirow}
\usepackage{color}
\usepackage{colortbl}
\usepackage[T1]{fontenc}
\usepackage{hyperref}

\usepackage[utf8]{inputenc}
\newcommand{\soloist}{\textsc{Soloist}}
\newcommand{\soloistteach}{\textsc{Soloist+Teach}}

\newcommand{\simpletod}{\textsc{SimpleTOD}}
\newcommand{\sptod}{\textsc{SGP-TOD}}
\newcommand{\ie}[0]{\emph{i.e., }}

\newcommand{\eg}[0]{\emph{e.g., }}

\newcommand{\etc}[0]{\emph{etc.}}
\newcommand{\aka}[0]{\emph{a.k.a. }}
\newcommand{\RN}[1]{%
	\textup{\lowercase\expandafter{\it \romannumeral#1}}%
}
\usepackage{microtype}

\usepackage{inconsolata}
\usepackage{makecell}
\usepackage{soul} 
\usepackage{arydshln}
\usepackage{booktabs}
\usepackage{multirow}
\usepackage{float}
\usepackage{color, xcolor} 
\definecolor{mypink1}{RGB}{255, 204, 204}
\definecolor{mygrey1}{RGB}{204,229,255}
\definecolor{myblue1}{RGB}{204, 255, 255}
\definecolor{mygreen1}{RGB}{204,255,204}
\definecolor{myyellow1}{RGB}{230,255,204}

\newcommand\blfootnote[1]{%
  \begingroup
  \renewcommand\thefootnote{}\footnote{#1}%
  \addtocounter{footnote}{-1}%
  \endgroup
}

\title{\sptod{}: Building Task Bots Effortlessly via \\Schema-Guided LLM Prompting}

\author{Xiaoying Zhang$^{1}$, Baolin Peng$^{2*}$, Kun Li$^{1}$, Jingyan Zhou$^{1}$, Helen Meng$^{1}$ \\
    $^{1}$The Chinese University of Hong Kong, Hong Kong\\
    $^{2}$Microsoft Research, Redmond\\
    \{zhangxy, kunli, jyzhou, hmmeng\}@se.cuhk.edu.hk  \\ 
  }

\begin{document}
\maketitle
\begin{abstract}

Building end-to-end task bots and maintaining their integration with new functionalities using minimal human efforts is a long-standing challenge in dialog research. 
Recently large language models (LLMs) have demonstrated exceptional proficiency in conversational engagement and adherence to instructions across various downstream tasks. In this work, we introduce \sptod{}, \b{S}chema-\b{G}uided \b{P}rompting for building \b{T}ask-\b{O}riented \b{D}ialog systems effortlessly based on LLMs. Utilizing the symbolic knowledge -- task schema, we instruct fixed LLMs to generate appropriate responses on novel tasks, circumventing the need for training data. Specifically, \sptod{} comprises three components: a LLM for engaging with users, a DST Prompter to aid the LLM with dialog state tracking, which is then used to retrieve database items, and a Policy Prompter to elicit proper responses adhering to the provided dialog policy. Experimental results on Multiwoz, RADDLE and STAR datasets show that our training-free strategy \sptod{}, without any task-specific data, yields state-of-the-art (SOTA) zero-shot performance, greatly surpasses the few-shot approaches. In a domain-extension setting, \sptod{} aptly adapts to new functionalities by merely adding supplementary schema rules. We make our code and data publicly available.\blfootnote{$^*$Currently at Tencent AI Lab. Work done at Microsoft Research. \hspace{3mm}}\footnote{\url{https://github.com/zhangxy-2019/sgp-tod} Preprint. Work in progress.}

\end{abstract}

\section{Introduction}


\begin{figure}[t]
\centering
\includegraphics[width=0.95\columnwidth]{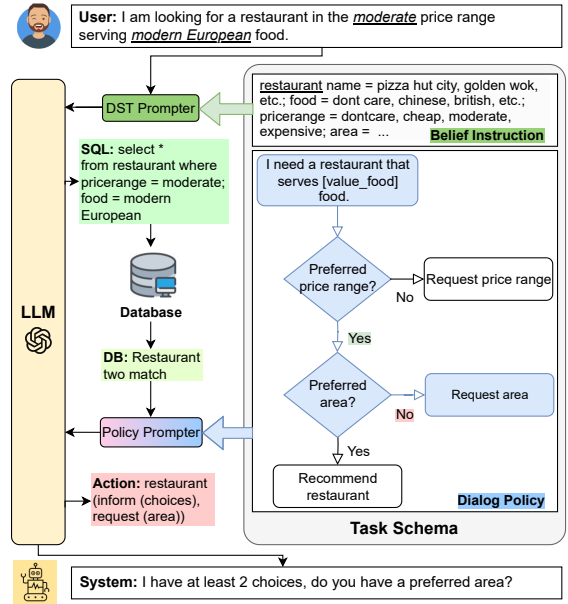}
\caption{The proposed \sptod{} is depicted with a dialog example, where the prompters integrate task schema (right) to assist the frozen LLM in generating an appropriate response (left).}
\label{fig:run_ex}
\end{figure}

Building task-oriented dialog (TOD) systems has been a long-standing challenge in artificial intelligence. 
The prevailing approach for creating task bots \cite{hosseini2020simple, peng2021soloist, sun2022mars} is to fine-tune pre-trained language models (PLMs), such as T5 \cite{2020t5} and GPT-2 \cite{radford2019language}. Despite their great success, developing and maintaining such task bots generally requires adequate annotated data and extensive fine-tuning/re-training. 
Recently, large Language Models (LLMs), such as ChatGPT\footnote{\url{https://chat.openai.com/}} and GPT-4 \cite{openai2023gpt4}, have revolutionized natural language processing (NLP) applications \cite{wang2023robustness, wei2022emergent}, owing to their remarkable conversational skills \cite{qin2023chatgpt}, instruction-following abilities~\cite{instructgpt2022} and zero-shot generalization capabilities~\cite{chowdhery2022palm, hu-etal-2022-context}. This raises a research question: can LLMs be effectively utilized for building task bots with minimum human efforts?

A contemporary study \cite{DBLP:journals/corr/abs-2304-06556} explores the potential of LLMs for rapidly building task bots via few-shot prompting, \aka in-context learning (ICL) paradigm \cite{NEURIPS2020_1457c0d6, madotto2021few}. Though demonstrably effective, the ICL performance is highly influenced by the quality of the in-context exemplars \cite{liu-etal-2022-makes, pmlr-v139-zhao21c, dong2023survey}, as they exhibit instability in conveying all requisite task instructions.

In this work, we introduce the symbolic knowledge \cite{Binder,DBLP:conf/nips/NyeTTL21} -- task schema into LLMs for creating task bots. Task schema \cite{mosig2020star, mehri2021schema} encompasses a concise symbolic representation of a task, supplying LLMs with a comprehensive blueprint. It comprises $(\RN{1})$ task-specific ontology containing all slots and their feasible values \cite{budzianowski2018large}; and $(\RN{2})$ a dialog flow explicitly outlining fundamental interaction patterns \cite{peng2021synergy}.
Specifically, we propose \sptod{} (as depicted in Figure \ref{fig:run_ex}), a schema-guided prompting method for rapidly building task bots. We integrate the predefined task schema and dialogue context into prompts through the use of two specifically-designed prompters, namely a DST Prompter and a Policy Prompter. Utilizing these prompters, we adeptly guide fixed LLMs to track dialogue states, retrieve database entries, and generate appropriate responses for novel tasks in a zero-shot manner, without the need for additional training or fine-tuning. 
By incorporating task-specific symbolic knowledge into LLMs, \sptod{} provides knowledge-based, coherent and human-like responses. Moreover, this training-free design empowers developers to flexibly prototype dialog systems on new tasks, while seamlessly extending their functionalities through modifying the task schema.

We conduct empirical experiments on two multi-domain datasets: Multiwoz 2.0 and 2.2 \cite{budzianowski2018large,zang-etal-2020-multiwoz}, and two single-domain/task datasets: RADDLE \cite{peng2021soloist} and STAR \cite{mosig2020star} in zero-shot scenarios. The results indicate that \sptod{}, \textit{employing merely task schema devoid of any training or fine-tuning}, substantially boosts the SOTA zero-shot results, markedly outperforming few-shot prompting/fine-tuning methods, and even attaining competitive results with full-shot fine-tuning approaches. In a domain-extension context, \sptod{} proficiently adapts to new functionalities \textit{by simply adding a handful of schema rules without necessitating further data collection}, significantly exceeding few-shot prompting/fine-tuning methods reinforced by machine teaching \cite{DBLP:conf/sigdial/WilliamsL17}.


In summary, our contributions are three-fold.

\begin{itemize}\setlength{\itemsep}{0pt}

\item We propose \sptod{}, a schema-guided prompting-only strategy that allows effortlessly building task bots based on LLMs.


\item We integrate symbolic knowledge -- task schema into LLMs, allowing them to generate schema-compliant responses and adaptively expand their functionalities to tackle new tasks by solely modifying the task schema.





\item We demonstrate the effectiveness of \sptod{} on Multiwoz, RADDLE, STAR datasets in zero-shot settings. \sptod{} attains SOTA zero-shot performance, notably surpassing few-shot prompting/fine-tuning methods and exhibiting favorable performance compared to full-shot fine-tuning methods.


\end{itemize}


\section{Related work}
\begin{table*}[!t]
\setlength\tabcolsep{4pt}
  \centering
  \begin{threeparttable}
  \fontsize{7}{7}
  \selectfont
    \begin{tabular}{lccccc}
    \toprule
    
    \multirow{2}{*}{Model}&\multirow{2}{*}{\texttt{Task}}&\multirow{2}{*}{\texttt{Schema types}}&
    \multicolumn{3}{c}{\texttt{Training strategy}}\cr\cmidrule(lr){4-6}
         &&&\texttt{Fine-tuning}&\texttt{Pre-training}&\texttt{Prompting} \cr
    \midrule
    \textsc{RobustSF} \cite{shah2019robust} &SF&\cellcolor{cyan!30}slot names/value examples&\Checkmark \\
    \midrule
    \textsc{TRADE} \cite{wu-etal-2019-transferable} &DST&\cellcolor{cyan!30}slot names/value examples&\Checkmark\\
    \textsc{ZSTL-SD} \cite{campagna-etal-2020-zero}&DST&\cellcolor{yellow!50}{ontology, dialog templates}&\Checkmark(+synthesized data) \\
    \cmidrule(lr){3-3}
    \multirow{2}{*}{\textsc{S-DST} \cite{rastogi2020towards}}&\multirow{2}{*}{DST} &\cellcolor{pink!70}{slot names/descriptions} &\multirow{2}{*}{\Checkmark} \cr &&\cellcolor{lightgray!50}{+service, intent names/descriptions} \\
    \cmidrule(lr){3-3}
    \textsc{T5DST} \cite{lin2021leveraging} &DST&\cellcolor{pink!70}{slot names/descriptions}&\Checkmark\\
    \textsc{TransferQA} \cite{lin2021zero} &DST&\cellcolor{cyan!30}slot names/value examples&&\Checkmark(QA tasks)\\
    \textsc{IC-DST} \cite{hu-etal-2022-context} &DST&\cellcolor{cyan!30}slot names/value examples&&&\Checkmark\\
    \textsc{SDM-DST} \cite{wang-etal-2022-slot} &DST&\cellcolor{cyan!30}slot names/value examples&\Checkmark\\
    \midrule
    \textsc{Bert+S} \cite{mosig2020star}&E2E policy&\cellcolor{lime!60}{system-side policy skeletons}&\Checkmark\\
    \textsc{SAM} \cite{mehri2021schema} &E2E policy&\cellcolor{lime!60}{user-aware policy skeletons}&\Checkmark\\
    \midrule
    \textsc{ZSDG} \cite{zhao2018zero} &E2E dialog&\cellcolor{yellow!50}{ontology, response templates}&\Checkmark \\
    \textsc{DAML} \cite{qian-yu-2019-domain} &E2E dialog&\cellcolor{yellow!50}{ontology, response templates}&\Checkmark \\
    \cmidrule(lr){3-3}
    \multirow{4}{*}{\textsc{AnyTOD} \cite{zhao2022anytod}}&\multirow{4}{*}{E2E dialog}&\cellcolor{lime!60}{policy programs} &\multirow{4}{*}{\Checkmark}&\multirow{4}{*}{\Checkmark(heterogeneous tasks)} \cr &&\cellcolor{cyan!30}+slot names/value examples \cr
    &&\cellcolor{pink!70}{+slot descriptions} \cr &&\cellcolor{orange!40}{+user action names/states/descriptions} \\
    \cmidrule(lr){3-3} 
    \multirow{2}{*}{\textsc{IT-LLM} \cite{DBLP:journals/corr/abs-2304-06556}} &\multirow{2}{*}{E2E dialog} & \cellcolor{cyan!30}slot names &&&\multirow{2}{*}{\Checkmark} \cr &&\cellcolor{pink!70}{+slot descriptions} \\
    \cmidrule(lr){3-3} 
    \multirow{2}{*}{\sptod{} (ours)}&\multirow{2}{*}{E2E dialog}&\cellcolor{lime!60}{user-aware policy skeletons} &&&\multirow{2}{*}{\Checkmark} \cr &&\cellcolor{cyan!30}(+slot names/value examples) \\
    \bottomrule  
    \end{tabular}
  \end{threeparttable}
  
  \caption{Zero-shot task-oriented dialog modeling. (Schema items enclosed in parentheses are required only when accessible.)}
  \label{tab:related}
  \vspace{-1mm}
\end{table*}

\paragraph{Zero-Shot Task-Oriented Dialog Modeling.} 
Zero-shot generalization is an essential yet challenging task in task-oriented dialog research. As shown in Table \ref{tab:related}, there are four main research directions: slot filling (SF), dialog state tracking (DST), end-to-end policy management (E2E policy) and end-to-end dialog generation (E2E dialog). 

In this paper, we focus on zero-shot end-to-end dialog modeling, including policy management and dialog generation. \citet{zhao2018zero, qian-yu-2019-domain} utilize ontology\footnote{Ontology is a structured representation of the back-end database, defining all slots and their possible values \cite{budzianowski2018large}.} and response templates to train the dialog model , enabling the discovery of shared dialog policies (\ie discourse-level patterns) between the source and target domains. To facilitate broader adaptation to previously unseen tasks or domains with diverse dialog policies, \citet{mosig2020star, mehri2021schema} implement task-specific policy skeletons, training dialog models to adhere to novel policies. Furthermore, \citet{zhao2022anytod} employs a neural language model (LM) for tracking dialog states and user actions using slot and action descriptions; subsequently, a policy program is executed to recommend the next system actions; ultimately, an LM generates the final system action and corresponding template response. Our \sptod{} diverges by necessitating fewer annotations, \ie we do not require state or action descriptions and integrate slot names and value examples into the task schema exclusively when accessible. Despite the effectiveness of previous approaches, they still require sufficient fine-tuning and annotated dialog corpora on source domains/tasks or heterogeneous tasks.

A concurrent study to ours is \citet{DBLP:journals/corr/abs-2304-06556}, which employs a prompting strategy to guide frozen LLMs in generating appropriate responses. Specifically, \citet{DBLP:journals/corr/abs-2304-06556} first tracks belief states based on the dialog history by utilizing slot descriptions as prompts, then retrieves database entries, and generates suitable actions and responses. In contrast, our \sptod{} differs in that: $(\RN{1})$ we employ slot names and value examples, rather than slot descriptions, as prompts to facilitate frozen LLMs in generating belief states, thereby reducing human effort; $(\RN{2})$ we offer a policy skeleton within the Policy Prompter to guide LLMs in producing responses that comply with the predefined dialog policy. In addition, experimental results indicate that \sptod{} substantially outperforms \textsc{IT-LLM} \cite{DBLP:journals/corr/abs-2304-06556}.


\paragraph{Leveraging LLMs for Dialog Tasks.}
LLMs \cite{NEURIPS2020_1457c0d6, DBLP:journals/corr/abs-2204-02311, chen2021evaluating,openai2023gpt4} have exhibited unparalleled mastery of natural language understanding, reasoning and generation \cite{wei2022emergent, bubeck2023sparks}. 

Three primary research directions have obtained substantial success in numerous dialog tasks by utilizing LLMs. 
$(\RN{1})$ Few-shot prompting \cite{NEURIPS2020_1457c0d6}, in which LLMs learn to execute new tasks by conditioning on a handful of in-context exemplars without training, has showcased remarkable performance in intent classification \cite{DBLP:conf/naacl/YuHZDPL21}, semantic parsing \cite{shin-van-durme-2022-shot}, dialog state tracking \cite{hu-etal-2022-context, xie-etal-2022-unifiedskg}, and response generation \cite{madotto2021few}. $(\RN{2})$ \citet{li-etal-2022-controllable, mehri-etal-2022-lad, dai2023auggpt} employ LLMs for data augmentation, \ie generating synthetic task-oriented dialogs to train smaller models for inference. 
$(\RN{3})$ Recently, several studies endeavor to support LLMs in specialized tasks by incorporating external knowledge \cite{peng2023check, liang2023taskmatrixai}. \citet{peng2023check} advocates for enhancing LLMs' responses with external knowledge and automated feedback to reduce hallucination in their responses. \citet{liang2023taskmatrixai} suggests connecting LLMs with millions of APIs to accomplish a variety of specialized tasks. Different from the aforementioned works, we aim to employ LLMs in constructing an end-to-end task-oriented dialog system in a zero-shot manner by using pre-defined task schema as prompts.



\begin{figure*}[!t]
\centering
\includegraphics[width=0.95\linewidth]{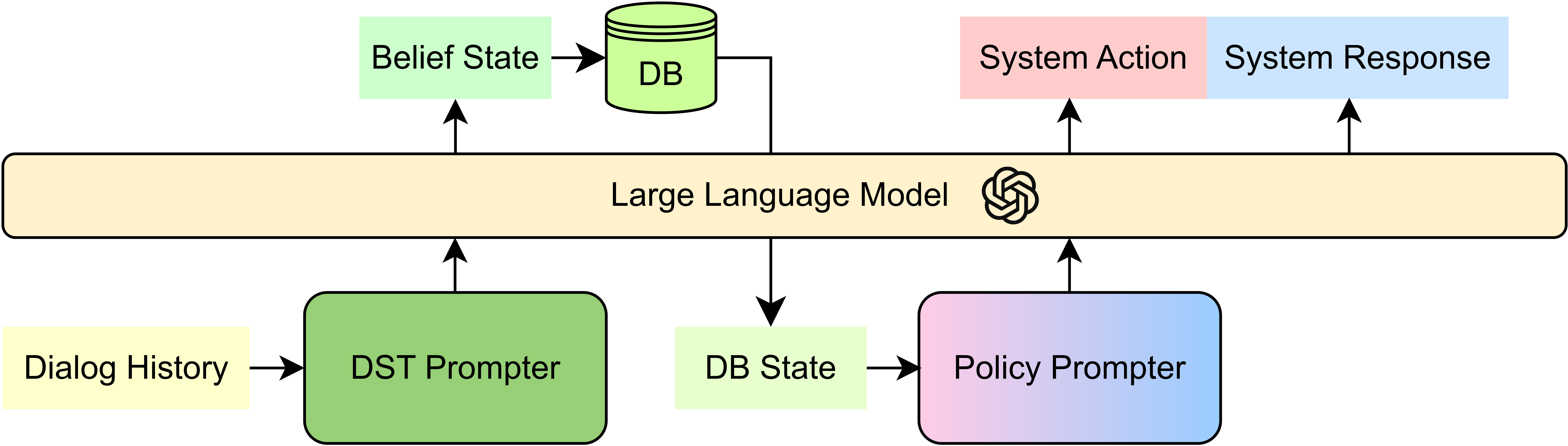}
\caption{Overview of \sptod{}. Detailed illustration with a dialog example is shown in Appendix \ref{sec:sptod_ex}.
}
\label{fig:operation_process}
\end{figure*}

\section{Methodology}

\subsection{Overview}
The overall architecture of the proposed \sptod{} (Figure \ref{fig:operation_process}) consists of three key components: $(\RN{1})$ a \textbf{LLM}, responsible for adhering to instructions, comprehending user queries, and generating coherent responses for user interaction; 
$(\RN{2})$ a \textbf{DST Prompter}, tasked with supporting the LLM in tracking dialogue states using belief instructions; 
$(\RN{3})$ a \textbf{Policy Prompter}, guiding the LLM to adhere to the predefined task policy for providing suitable system actions and responses.

As depicted in Figure \ref{fig:operation_process}, at each dialog turn $t$, the end-to-end generation task is systematically divided into three subsequent sub-tasks:
$(\RN{1})$ \textbf{Belief State Prediction} -- given the dialog history up to current dialog turn $\boldsymbol{h}_{t}$, which is a sequence of utterances alternating between the user and the system $\boldsymbol{h}_{t} = [u_1, r_1, u_2, r_2, \dots, u_t]$ (where $u$ and $r$ denote user and system utterances, respectively), the DST Prompter embeds the belief instruction $\boldsymbol{BI}$ to direct the frozen LLM (parameterized by $\boldsymbol{\theta}$) in generating a belief state $\boldsymbol{b}_{t}$ (Equation \ref{eqa:x1}). The belief state is then used to query a database and obtain the database (DB) state $\boldsymbol{c}_{t}$ (Equation \ref{eqa:x2}).
$(\RN{2})$ \textbf{System Action Determination} --
the Policy Prompter incorporates a policy skeleton $\boldsymbol{PS}$, assisting the language model in generating a system action $\boldsymbol{a}_{t}$, $\boldsymbol{a}_{t}$ based on $\boldsymbol{h}_{t}$, $\boldsymbol{b}_{t}$, and $\boldsymbol{c}_{t}$ (Equation \ref{eqa:x3}).
$(\RN{3})$ \textbf{Dialog Response Generation} -- grounded in the dialog history $\boldsymbol{h}_{t}$, belief state $\boldsymbol{b}_{t}$, DB state $\boldsymbol{c}_{t}$, system action $\boldsymbol{a}_{t}$, the Policy Prompter aids the LLM in generating a delexicalized response by providing the policy skeleton $\boldsymbol{PS}$ (Equation \ref{eqa:x4}).  Ultimately, the delexicalized dialog system is automatically post-processed to generate system response in natural language.
\begin{align}
    \boldsymbol{b}_{t} = & \boldsymbol{LLM}_{\boldsymbol{\theta}} ( \boldsymbol{h}_{t}, \boldsymbol{BI})
    \label{eqa:x1} \\
    \boldsymbol{c}_{t} = & \boldsymbol{DB}(\boldsymbol{b}_{t}) 
    \label{eqa:x2} \\
    \boldsymbol{a}_{t} = & \boldsymbol{LLM}_{\boldsymbol{\theta}} ( \boldsymbol{h}_{t}, \boldsymbol{b}_{t}, \boldsymbol{c}_{t}, \boldsymbol{PS}) 
    \label{eqa:x3} \\
    \boldsymbol{r}_{t} = & \boldsymbol{LLM}_{\boldsymbol{\theta}} ( \boldsymbol{h}_{t}, \boldsymbol{b}_{t}, \boldsymbol{c}_{t}, \boldsymbol{a}_{t}, \boldsymbol{PS}) 
    \label{eqa:x4} 
\end{align}


\subsection{LLM}

A LLM is responsible for following task-specific instructions and generating appropriate responses.

Many off-the-shelf LLMs, \eg ChatGPT, Codex \cite{chen2021evaluating}, are pre-trained on massive corpora of text data and/or code data. In addition, they are trained to follow instructions in the prompts~\cite{instructgpt2022} and provide pertinent responses. Exhibiting remarkable proficiencies in natural language processing, instruction compliance, and zero-shot generalization across diverse downstream dialog tasks, these LLMs serve as valuable foundation models for our approach.

\subsection{DST Prompter}
\label{sec:dst_prop}

Given the dialog history $\boldsymbol{h}_{t}$, the DST prompter aims to guide the LLM in predicting the belief state $\boldsymbol{b}_{t}$ at each turn $t$, using the belief instruction $\boldsymbol{BI}$. The belief state is defined as the concatenation of the domain/task (\ie user intent) $\boldsymbol{d}_{t}$ and a set of slot-value pairs $\left\{(\boldsymbol{s}_{t}^{1}, \boldsymbol{v}^{1}_{t}); \dots ; (\boldsymbol{s}_{t}^{n_{t}}, \boldsymbol{v}_{t}^{n_{t}}) \right\}$:
\begin{equation}
     \boldsymbol{b}_{t} = \boldsymbol{d}_{t}, \left\{(\boldsymbol{s}_{t}^{i}, \boldsymbol{v}_{t}^{i}); i = 1, \dots, n_{t} \right\}
    \label{eqa:x6} 
\end{equation}
\noindent where $n_{t}$ is the total number of pairs in the set.

\begin{figure}[t]
\centering
\includegraphics[width=0.95\columnwidth]{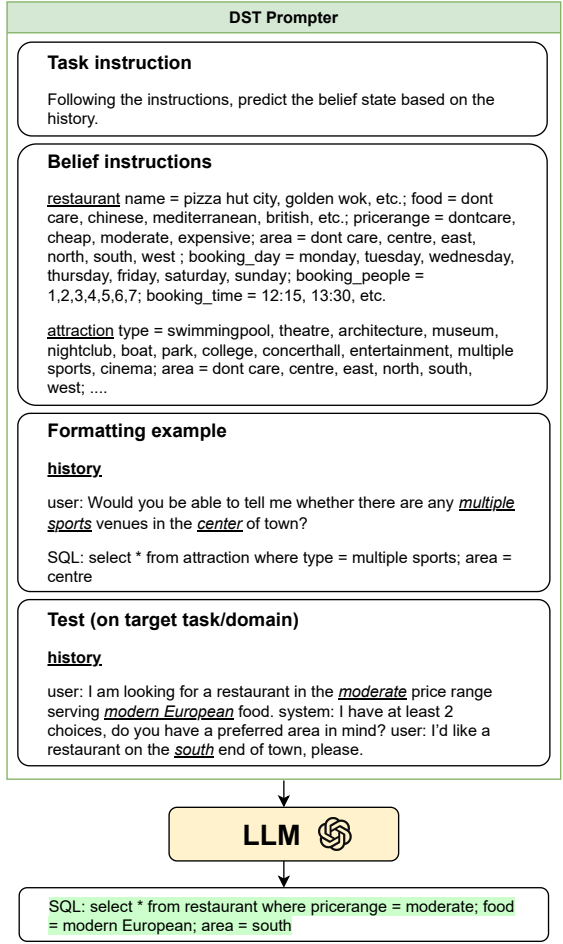}
\caption{Illustration of belief state prediction utilizing DST Prompter. The \colorbox{mygreen1}{predicted belief state} is highlighted.}

\label{fig:belief_instruct_demo}
\end{figure}

As shown in Figure \ref{fig:belief_instruct_demo}, the proposed DST prompter contains four parts: $(\RN{1})$ a \textit{task instruction} that offers general guidance on belief state prediction;\footnote{We assess several task instructions written by different authors, yielding minor performance disparities.} $(\RN{2})$ \textit{belief instructions} $\boldsymbol{BI}$ of all domains/tasks; $(\RN{3})$ a \textit{formatting example} illustrating the anticipated output format to direct the LLM, in addition, we follow \citet{hu-etal-2022-context} and adopt SQL state to represent the dialog state $\boldsymbol{b}_{t}$\footnote{SQL: select * from $\boldsymbol{d}_{t}$ where $\boldsymbol{s}_{t}^{1} = \boldsymbol{v}_{t}^{1}; \dots; \boldsymbol{s}_{t}^{n_{t}}=\boldsymbol{v}_{t}^{n_{t}}$.}; and $(\RN{4})$ the \textit{test input}, \ie the given dialog history $\boldsymbol{h}_{t}$. Since the prompt is fixed and no labeled data from the target task or domain is used, we refer to this setting as "zero-shot", following \citet{wang2022super}.

\paragraph{Belief Instruction.}
For each task/domain, the belief instruction contains the task/domain name, all potential slot names, and their possible values (Figure \ref{fig:belief_instruct_demo}). Regarding categorical slots, such as the "price range" in the restaurant domain, all plausible values are included, \ie "don't care", "cheap", "moderate", and "expensive"; whereas, for non-categorical slots, such as "name", only a select few value examples are injected, \eg Pizza Hut City, Golden Wok, etc.\footnote{We assess belief instructions with diverse slot value examples, revealing minor performance variations.} Detailed belief instructions for all tasks/domains can be found in Appendix \ref{sec:dst_bi}.

\begin{figure}[t]
\centering
\includegraphics[width=0.95\columnwidth]{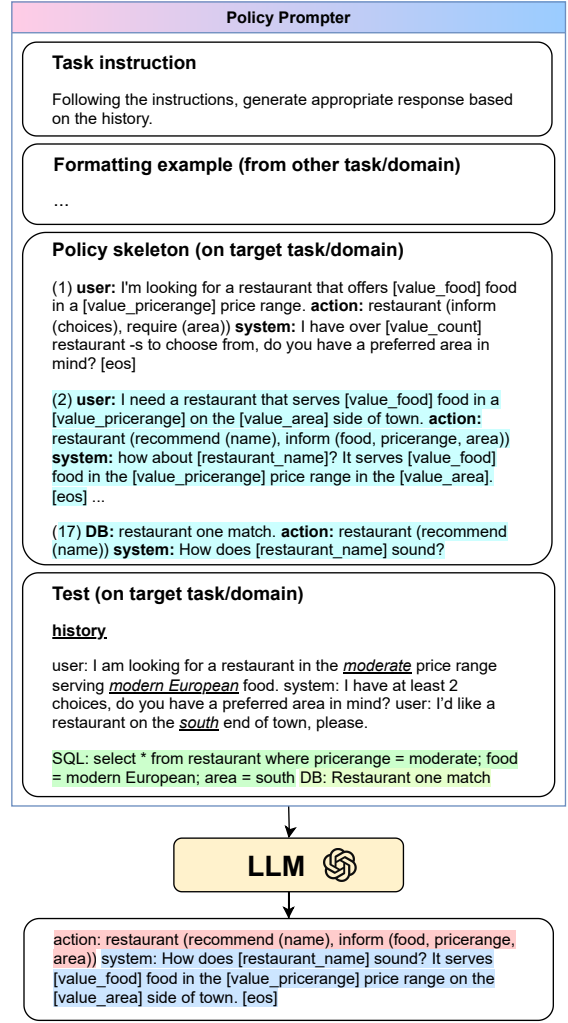}
\caption{Illustration of system action determination and response generation employing the Policy Prompter. The \colorbox{myblue1}{pertinent template turn}, previous \colorbox{mygreen1}{predicted belief state}, \colorbox{myyellow1}{retrieved DB state} within the input, alongside the \colorbox{mypink1}{generated system action} and \colorbox{mygrey1}{generated response} in the output are accentuated.}
\label{fig:policy_instruct_demo}
\end{figure}

\subsection{Policy Prompter}
Dialog policy, governing the behavior of task bots, plays a crucial role in task-oriented dialogs \cite{zhao2022anytod}. To represent the dialog policy for a given task, we utilize a \textit{policy skeleton}, which delineates interaction patterns and encompasses business logic in the form of template dialog flows~\cite{peng2021synergy}. The Policy Prompter is devised to guide the static LLM in adhering to the policy skeleton $\boldsymbol{PS}$, enabling the sequential generation of appropriate system actions $\boldsymbol{a}_{t}$ and responses $\boldsymbol{r}_{t}$.

Analogous to the DST Prompter, the Policy Prompter (Figure \ref{fig:policy_instruct_demo}) comprises four components: $(\RN{1})$ a \textit{task instruction}; $(\RN{2})$ a \textit{formatting example} originating from another task/domain, entailing a partial policy skeleton and its corresponding dialogue turn exemplar (found in Appendix \ref{sec:format_ex}); $(\RN{3})$ a \textit{policy skeleton} for the previously predicted domain/task; and $(\RN{4})$ the \textit{test input}, \ie the dialog history $\boldsymbol{h}_{t}$, generated belief state $\boldsymbol{b}_{t}$, and obtained DB state $\boldsymbol{c}_{t}$.


\paragraph{Policy Skeleton.}

Given that user behaviors and DB results jointly determine system actions and responses, policy skeleton is designed to cover all fundamental user behaviors and characteristic DB results, along with their corresponding system actions and responses.\footnote{It should be noted that we do not enumerate every conceivable combination of user behaviors or potential database outcomes, as schema engineering is not the primary focus of this study.)} 
Considering the infeasibility of developing a multi-task/domain policy skeleton for every possible combination of tasks and domains, we opt to develop a distinct policy skeleton tailored to each specific task and domain.

Following \citet{mehri2021schema}, our strategy converts the established dialog policy into a series of template dialog turns $\mathcal{X}$ that are logically arranged and concentrate on task completion:
\begin{equation}
\begin{aligned}
    \mathcal{X} = &\left\{\boldsymbol{x}_{i}\right\}_{i=1}^{N}, \\
    \boldsymbol{x}_{i} = & (u^{i},a^{i},r^{i}) or (c^{i},a^{i},r^{i})
\end{aligned}
\end{equation}
\noindent where $\boldsymbol{x}_{i}$ is a template dialog turn, which contains a user utterance $u^{i}$ or a DB state $c^{i}$, matching system action $a^{i}$, and system response $r^{i}$. $N$ denotes the total number of template turns within the policy skeleton (around 10-20 template turns depending on the task complexity). In order to equip the frozen LLM with additional capabilities or modify current ones, we only need insert, amend, or eliminate a few template turns within the policy skeleton.

\section{Experiments}






\subsection{Experimental Setup}
\label{sec:ex_set}
\paragraph{Datasets.} 
We validate the effectiveness of \sptod{} on the following dialog datasets: 

\begin{itemize}\setlength{\itemsep}{0pt}
\item \texttt{Multiwoz 2.0} \cite{budzianowski2018large} is a \textbf{multi-domain} task-oriented dataset, which contains 8,438/1,000/1,000 dialogs for training/validating/testing, spanning seven domains: restaurant, attraction, train, hotel, taxi, police, and hospital. \texttt{Multiwoz 2.0} is annotated with belief states and system actions.

\item \texttt{Multiwoz 2.2} \cite{zang-etal-2020-multiwoz} is a improved version of \texttt{Multiwoz 2.0}, encompassing refined belief state annotations, slot descriptions, user action annotations, \etc 

\item \texttt{RADDLE} \cite{peng2021soloist,DBLP:conf/acl/PengLZZLG20} consists of four \textbf{single-domain} dialog datasets derived from \texttt{Multiwoz 2.0} (\ie restaurant, train, hotel, attraction), reorganized by \citet{peng2021soloist}.
Each corpus contains 50/50/100-200 dialogs for training/validating/testing.

\item \texttt{STAR} \cite{mosig2020star} includes 24 tasks in 13 domains (\eg "apartment" domain comprises "apartment-search" and "apartment-schedule"), requiring the dialog model to conform to the provided task schema. 
We use 2,688 single-task dialogs from the corpus, which follow a "happy path", \ie the user is not instructed to execute any action exceeding the schema's expectations. 
Without additional annotations, \texttt{STAR} only provides a flow chart diagram that outlines the dialog policy for each task.
\end{itemize}

\paragraph{Automatic Evaluation Metrics.} 




We evaluate the end-to-end dialog generation performance using the same metrics as those listed in \citet{budzianowski2018large}: 
$(\RN{1})$ $\mathtt{Inform}(\%)$ assesses whether the agent returns an acceptable entity. 
$(\RN{2})$ $\mathtt{Success}(\%)$ determines if the agent appropriately responds to each attribute request.
$(\RN{3})$ $\mathtt{BLEU}(\%)$ \cite{papineni-etal-2002-bleu} measures the word overlap of the generated response against the human response in the corpus. 
$(\RN{4})$ $\mathtt{Combined}(\%)$ judges the overall quality, which is defined as $\mathtt{Combined}$ = ($\mathtt{Inform}$ + $\mathtt{Success}$) $\times$ 0.5 + $\mathtt{BLEU}$. Additionally, we utilize $\mathtt{BERTScore}(\%)$ \cite{bert-score}, which focuses on computing semantic similarity between the generated responses and the ground truth, and correlates better with human judgments.

Following \citet{mehri2021schema}, we perform the next action prediction task on \texttt{STAR}, which predicts next system action based on the dialog history. 
Since the system actions and deterministic response templates are mapped one to one in \texttt{STAR} corpus, we believe the end-to-end next action prediction task falls within end-to-end dialog modeling, following \citet{mosig2020star, mehri2021schema}.
In addition, we report the results using weighted $\mathtt{F1score}(\%)$ and mean $\mathtt{accuracy}(\%)$.

\begin{table*}[!t]
\setlength\tabcolsep{4pt}
  \centering
  \begin{threeparttable}
  \fontsize{8}{9}
  \selectfont
    \begin{tabular}{lcccccccc}
    \toprule
    \multirow{2}{*}{Model}&
    \multicolumn{4}{c}{\texttt{Multiwoz 2.0}}
     &\multicolumn{4}{c}{\texttt{Multiwoz 2.2}}\cr\cmidrule(lr){2-5} \cmidrule(lr){6-9}
     &$\mathtt{Inform}$&$\mathtt{Success}$&$\mathtt{BLEU}$&$\mathtt{Combined}$&$\mathtt{Inform}$&$\mathtt{Success}$&$\mathtt{BLEU}$&$\mathtt{Combined}$ \cr
    \midrule
    \multicolumn{9}{l}{\textit{Full-shot fine-tuning (with 8.4k+ training dialogs):}} \\
     \textsc{DAMD} \cite{zhang2020task} &76.33& 60.40 &16.60 &84.97 &-&-&-&-\\
 \simpletod{} \cite{hosseini2020simple}&84.40&70.10&15.01&92.26 &-&-&-&-\\
    \textsc{UBAR} \cite{yang2021ubar}&85.10&71.02&16.21&94.27 &-&-&-&-\\
    \textsc{MinTL} \cite{lin-etal-2020-mintl}&80.04&72.71&19.11&95.49 &-&-&-&- \\
    \soloist{} \cite{peng2021soloist}&85.50& 72.90 &16.54 &95.74&81.70 &67.10 &13.60&88.00\\
    \textsc{PPTOD} \cite{su2021multitask} &89.20 &79.40 &18.62 &102.92 &-&-&-&-\\
    \textsc{Mars} \cite{sun2022mars} &88.90 &78.00 &19.90 &103.35&88.90 &78.00& 19.60&103.05\\
    \midrule
    \multicolumn{9}{l}{\textit{Zero-shot transfer method (pre-trained on SGD):}} \\
     \textsc{AnyTOD-XXL}\cite{zhao2022anytod} &-&-&-&-&73.90 &24.40 &3.40&52.55\\
    \multicolumn{9}{l}{\textit{Few-shot prompting:}} \\
     \textsc{IT-LLM-ChatGPT-fs} \cite{DBLP:journals/corr/abs-2304-06556} &-&-&-&-&-&20.00&7.17&-\\
     \textsc{Few-Shot-ChatGPT} &44.74&24.32&7.88&42.41&45.40&24.50&7.72&42.67\\

    \multicolumn{9}{l}{\textit{Zero-shot prompting:}} \\
    \textsc{IT-LLM-ChatGPT-zs} \cite{DBLP:journals/corr/abs-2304-06556} &-&-&-&-&-&15.00&3.58&-\\
    \textsc{SGP-TOD-ChatGPT} (Ours) &64.56&54.05&7.17&66.48&64.70&54.70&6.96&66.66\\
    \textsc{SGP-TOD-Codex} (Ours) &71.67&52.55&7.91&70.02&75.50&52.30&6.62&70.53\\
    \textsc{SGP-TOD-GPT3.5} (Ours) &\textbf{83.88}&\textbf{69.87}&\textbf{9.09}&\textbf{85.97}&\textbf{82.00}&\textbf{72.50}&\textbf{9.22}&\textbf{86.47}\\
    \bottomrule  
    \end{tabular}
  \end{threeparttable}
  
  \caption{End-to-end dialog generation evaluation on \texttt{Multiwoz}. Results of \soloist{}, \textsc{Mars}, \textsc{AnyTOD-XXL} on \texttt{Multiwoz 2.2} are cited from \citet{zhao2022anytod}. Results of \textsc{IT-LLM-ChatGPT} are cited from \citet{DBLP:journals/corr/abs-2304-06556}. Other results of the full-shot fine-tuning methods are cited from \citet{he2022galaxy} and \citet{sun2022mars}. (We do not report the performance of \textsc{AnyTOD-XXL} and \textsc{IT-LLM-ChatGPT} on \texttt{Multiwoz 2.0}, because their code is not publicly availvable. Difference in mean is significant with p<0.01.)}
  \label{tab:multi-domain}
  \vspace{-1mm}
\end{table*}

\begin{table*}[!t]
\setlength\tabcolsep{3.5pt}
\fontsize{7}{7}
\selectfont
  \centering
  \begin{threeparttable}
    \begin{tabular}{lcccccccccccccccc}
    \toprule  
    \multirow{2}{*}{Model} &
    \multicolumn{4}{c}{\texttt{Attraction}}&\multicolumn{4}{c}{\texttt{Train}}&\multicolumn{4}{c}{\texttt{Hotel}}&\multicolumn{4}{c}{\texttt{Restaurant}}\cr  
      \cmidrule(lr){2-5} \cmidrule(lr){6-9} \cmidrule(lr){10-13} \cmidrule(lr){14-17}
      &$\mathtt{Info.} $&$\mathtt{Succ.}$&$\mathtt{BLEU}$&$\mathtt{Comb.}$&$\mathtt{Info.}$&$\mathtt{Succ.}$&$\mathtt{BLEU}$&$\mathtt{Comb.}$&$\mathtt{Info.}$&$\mathtt{Succ.}$&$\mathtt{BLEU}$&$\mathtt{Comb.}$&$\mathtt{Info.}$&$\mathtt{Succ.}$ &$\mathtt{BLEU}$&$\mathtt{Comb.}$ \cr 
    \midrule
    \multicolumn{17}{l}{\textit{Few-shot fine-tuning (with 50 training dialogs):}}\cr  
     \textsc{DAMD} &70.00&15.00&6.90&49.40&75.00&39.50&6.20& 63.45&62.50&20.50&7.60&49.10&68.00&19.50&10.50&54.50\cr
     \simpletod{}&65.66&46.97&5.85&62.17&59.00&44.00&7.07&58.57&62.50&40.00&7.70&58.95&75.50&
     44.50&11.00&71.00\cr
    \soloist{} &86.00&65.00&\textbf{12.90}& 88.40 &80.81&64.65&\textbf{9.96}& 82.69&74.50&43.50&\textbf{8.12}&67.12&81.00&55.50&12.80 &81.50\cr
    \midrule
    \multicolumn{17}{l}{\textit{Few-shot prompting:}}\cr
    \textsc{Few-Shot-ChatGPT} & 75.00&67.00&8.22&79.23&79.80&65.66&8.12&80.85&51.00&26.50&5.80&44.55&80.00&55.50&7.71&75.46 \\
    \multicolumn{17}{l}{\textit{Zero-shot prompting:}}\cr  
     \textsc{SGP-TOD-ChatGPT} & 95.00 & \textbf{94.00} & 7.13 &101.63 &76.77 & 74.24 & 6.75 & 82.26 &76.50 & 57.00 & 5.16 & 71.91 & 90.00&82.50&6.72 &92.97\\
     \textsc{SGP-TOD-Codex}  & \textbf{98.00} & 93.00 & 10.45 &\textbf{105.95} &78.79 & 70.20 & 8.56 & 83.06 &\textbf{83.50} & 69.50 & 7.86 & \textbf{84.36} & 91.00&\textbf{85.00}&10.50 &98.50\\
     \textsc{SGP-TOD-GPT3.5} &96.00&93.00&9.53&104.03&\textbf{82.83}&\textbf{77.27}&8.72&\textbf{88.77}&82.50&\textbf{71.50}&7.05&84.05&\textbf{91.50}&84.00&\textbf{12.90}&\textbf{100.65} \\
    \bottomrule  
    \end{tabular}
  \end{threeparttable}
  \caption{End-to-end dialog generation evaluation on \texttt{RADDLE}. The few-shot fine-tuning results are cited from \citet{peng2021soloist}. (Difference in mean is significant with p<0.01.)
  }
  \label{tab:results_4task}
\end{table*}

\paragraph{Comparison Methods.} 
 
 We evaluate the zero-shot performance of the proposed \textsc{SGP-TOD} by comparing it to SOTA zero-shot transfer methods and zero-shot/few-shot prompting strategies. (We report the mean results of three different runs.)

\noindent \textbf{Zero-shot transfer methods:}
\begin{itemize}\setlength{\itemsep}{0pt}
    \item \textsc{BERT+S} \cite{mosig2020star} is a schema-guided method that augments a BERT-base classifier \cite{DBLP:conf/naacl/DevlinCLT19} with a provided system-side schema to predict the next system action.
    \item \textsc{SAM} \cite{mehri2021schema} represents a schema-guided model based on BERT-base, which aligns the dialog context to a user-aware schema to predict the next system action.
    \item \textsc{AnyTOD-XXL} \cite{zhao2022anytod} adopts a neural LM to track dialog states and user actions utilizing slot and action descriptions. Then a program that outlines a predefined task policy is executed to recommend appropriate system actions. 
    Upon considering these system actions, a LM generates the ultimate system action and formulates the corresponding template response using the approach proposed by \citet{DBLP:conf/emnlp/KaleR20}.
    \textsc{AnyTOD-XXL} is implemented on T5-XXL \cite{roberts2022scaling}.

\end{itemize}

\noindent \textbf{Prompting methods:}
\begin{itemize}\setlength{\itemsep}{0pt}
    \item \textsc{IT-LLM-ChatGPT} \cite{DBLP:journals/corr/abs-2304-06556} is a prompting approach based on ChatGPT that leverages the dialog context and manually-crafted slot descriptions as the prompt, to track dialog states, fetch DB entries, and produce responses. 
    \textsc{IT-LLM-ChatGPT} incorporates one formatting example in zero-shot scenarios, denoted as \textsc{IT-LLM-ChatGPT-zs}, and integrates four task-specific examples retrieved from the training corpus in few-shot settings, \ie \textsc{IT-LLM-ChatGPT-fs}.
    
    \item \textsc{Few-Shot-ChatGPT} is a few-shot prompting strategy implemented on ChatGPT, where we use a few (\ie $\boldsymbol{k}$) dialog turns, randomly sampled from the training corpus to instruct ChatGPT on task execution. Upon evaluating various configurations of $\boldsymbol{k}$, the optimal results manifest with 15 on \texttt{Multiwoz} (2.0 and 2.2), and 10 on \texttt{RADDLE}, exhibiting no further substantial enhancements.
   
    \item \textsc{SGP-TOD} (Ours) is a schema-guided prompting strategy, which is compatible with any off-the-shelf LLMs. In this paper, we employ ChatGPT ("gpt-3.5-turbo"), GPT-3.5 ("text-davinci-003") and Codex ("code-davinci-002") as the fixed LLMs. Following the zero-shot scenario in \citet{wang2022super}, we insert one formatting example from different tasks (fixed through the experimental procedure) into the prompt. More implementation details are provided in Appendix \ref{sec:imple_details}.  
    
\end{itemize}
\begin{figure*}[!t]
\centering
\subfigure[Task transfer]{
\begin{minipage}[t]{0.5\linewidth}
  \centering
  \includegraphics[width=3.0in]{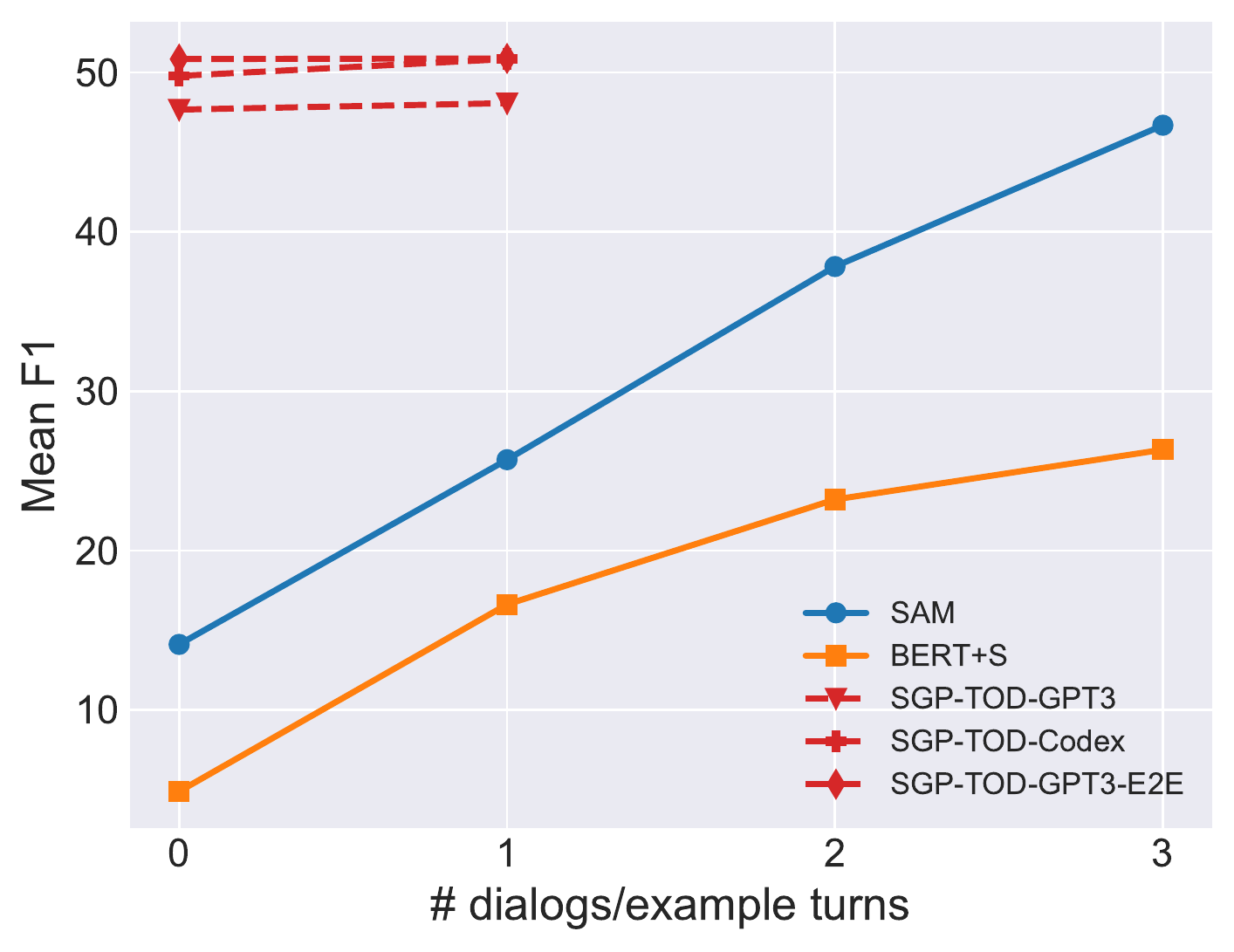}
  \label{fig:trans_f1}
\end{minipage}%
}%
\subfigure[Domain transfer]{
\begin{minipage}[t]{0.5\linewidth}
  \centering
  \includegraphics[width=3.0in]{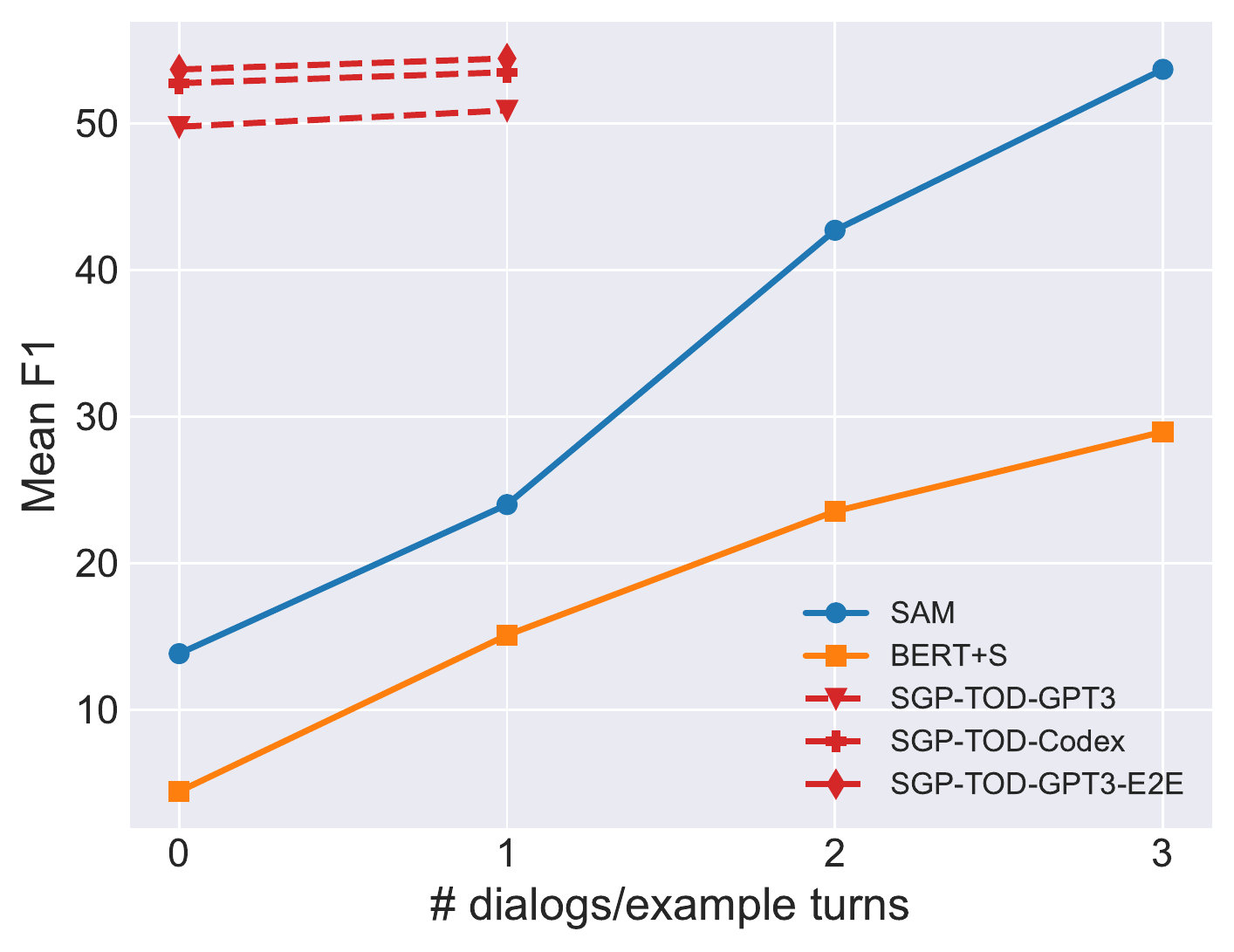}
  \label{fig:domain_f1}
\end{minipage}%
}%
\caption{Zero-shot evaluation results on \texttt{STAR} with different number of training dialogs (1, 10, 100, 1,000) / formatting example turns (1, 10) randomly sampled from the source domain/tasks. (Note the numbers are represented in logarithmic form with base 10.)}
\label{zero-shot-star}
\end{figure*}

\subsection{End-to-End Evaluation on \texttt{Multiwoz}}
\paragraph{Setup.}
\textsc{AnyTOD-XXL} is pre-trained on \texttt{SGD} dataset \cite{DBLP:conf/aaai/RastogiZSGK20}\footnote{\texttt{SGD} is a large-scale schema-guided multi-domain dialog dataset, spanning 45 domains.} then evaluated on \texttt{Multiwoz} in a zero-shot manner. 

\paragraph{Results.}
We present the evaluation results in multi-domain contexts on \texttt{Multiwoz} in Table \ref{tab:multi-domain}. In addition to the aforementioned methods, we include the results of SOTA full-shot fine-tuning approaches to facilitate a more comprehensive comparison.
\sptod{} obtains SOTA zero-shot performance, substantially outperforming few-shot approaches across all evaluation metrics, while even exhibiting competitive results in comparison to full-shot fine-tuning methods concerning $\mathtt{Success}$ and $\mathtt{Inform}$.
This confirms the effectiveness of integrating the task schema with the LLMs' proficient language processing capabilities and instruction-following abilities.

\paragraph{\textit{Comparison with Prompting Methods.}}
\textsc{SGP-TOD-ChatGPT} distinctly surpasses the zero-shot prompting approach \textsc{IT-LLM-ChatGPT-zs} with respect to $\mathtt{Success}$ (surpassing by $40\%$) and $\mathtt{BLEU}$ (exceeding by $3\%$). Moreover, \textsc{SGP-TOD-ChatGPT}, \textit{without requiring task-specific data}, considerably outperforms the few-shot prompting methods, \ie \textsc{IT-LLM-ChatGPT-fs} and \textsc{Few-Shot-ChatGPT} (\eg about 30 points improvement over $\mathtt{Success}$). 
This suggests that providing explicit, concise and comprehensive task instructions via task schema is preferable to imparting implicit task guidance through the selected dialog turns from the training corpus.


\paragraph{\textit{Comparison with Zero-Shot Transfer Methods.}}
Our \sptod{} demonstrates a substantial advantage over the \textsc{AnyTOD-XXL}, which necessitates extensive pre-training and supplementary annotations, \eg slot/action descriptions, over all the metrics. 
This exemplifies the potency of \sptod{}, which markedly reduces the necessity for human labor and computational resources.

\paragraph{\textit{Comparison with Full-Shot Fine-Tuning Methods.}} 
\textsc{SGP-TOD}, \textit{without training on any task-specific data}, exhibits competitive performance over $\mathtt{Inform}$ and $\mathtt{Success}$. 
This notable performance can be attributed to: $(\RN{1})$ the utilization of manually-crafted belief instructions and policy skeletons (derived from task schema) that serve as a good starting point for injecting symbolic knowledge for task completion; and $(\RN{2})$ LLM's exceptional ability to follow instructions and generate coherent texts. 
The inferior performance in $\mathtt{BLEU}$ is due to a lack of linguistic variations of the template utterances, which is acceptable considering the trade-off between human effort and efficacy
\paragraph{\textit{Impact of Different LLMs.}}
We find that \textsc{SGP-TOD-GPT3.5} performs better than \textsc{SGP-TOD-Codex} and \textsc{SGP-TOD-ChatGPT}. 

\subsection{End-to-End Evaluation on \texttt{RADDLE}}

\paragraph{Results.}
Table \ref{tab:results_4task} reports the end-to-end performance in single-domain settings on \texttt{RADDLE}. On all four dialog tasks, \sptod{} achieves considerably  superior performance over $\mathtt{Combined}$ compared to few-shot prompting and fine-tuning approaches. This indicates that even within a single-domain context (\ie encompassing a notably simpler task policy), furnishing explicit task instructions employing task schema remains preferable to rendering implicit task directions within dialog turns.

\paragraph{\textit{Comparison with Prompting Method.}}
Our \textsc{SGP-TOD} significantly surpasses \textsc{Few-Shot-ChatGPT} across all the metrics, aligning with the results observed in multi-domain contexts and further substantiating its academic significance.


\paragraph{\textit{Comparison with Few-Shot Fine-Tuning Methods.}}
\textsc{SGP-TOD} exhibits a zero-shot performance that consistently surpasses few-shot fine-tuning approaches in regard to $\mathtt{Inform}$, $\mathtt{Success}$, and $\mathtt{Combined}$, while remaining competitive in terms of $\mathtt{BLEU}$. The exceptional zero-shot performance of \textsc{SGP-TOD} is noteworthy, considering that \soloist{} is a powerful dialog model pre-trained on a number of heterogeneous dialog corpora and subsequently fine-tuned on \texttt{RADDLE}.

\subsection{End-to-End Evaluation on \texttt{STAR}}

\begin{table}[!t]
\setlength\tabcolsep{5pt}
  \centering
  \begin{threeparttable}
  \fontsize{7.5}{8}
  \selectfont
    \begin{tabular}{lccccc}
    \toprule
    \multirow{2}{*}{Model}&
    \multicolumn{2}{c}{\texttt{Task transfer}} & \multicolumn{2}{c}{\texttt{Domain transfer}}\cr
    \cmidrule(lr){2-3} \cmidrule(lr){4-5}
    &$\mathtt{F1}$&$\mathtt{Accuracy}$&$\mathtt{F1}$&$\mathtt{Accuracy}$ \cr
    \midrule
    \multicolumn{5}{l}{\textit{Zero-shot transfer}}\cr  
    \multicolumn{5}{l}{\textit{(leave-one fune-tuning with 2.5k training dialogs):}}\cr  
    \textsc{BERT+S}&24.25&24.89&25.70&28.56\\
    \textsc{SAM}&49.82&\textbf{51.30}&\textbf{55.91}&\textbf{57.92}\\
    \multicolumn{5}{l}{\textit{Zero-shot prompting:}}\cr  
    \textsc{SGP-TOD-Codex-INI}&45.18&47.99&47.21&49.97 \\
    \textsc{SGP-TOD-GPT3.5}&47.67&48.27&49.76&50.39\\
    \textsc{SGP-TOD-Codex}&49.78&51.01&52.72&53.66\\
    \textsc{SGP-TOD-GPT3.5-E2E}&\textbf{50.84}&50.74&53.50&53.21\\
    \bottomrule  
    \end{tabular}
  \end{threeparttable}
  \caption{Zero-shot end-to-end next action prediction evaluation on \texttt{STAR}. (Difference in mean is significant with p<0.01.)}
  \label{tab:star}
  \vspace{-1mm}
\end{table}

\paragraph{Setup.}
As mentioned in Section \ref{sec:ex_set}, we conduct next action prediction task on \texttt{STAR}, following \citet{mehri2021schema}. 
\textsc{BERT+S}, \textsc{SAM} are assessed in the leave-one fine-tuning scenario, where the models are fine-tuned on source tasks/domains in \texttt{STAR} then zero-shot on the held-out task/domain.\footnote{\textsc{AnyTOD-XXL} requires additional annotations, \eg belief descriptions, which makes it not suitable for \texttt{STAR}.} 
\sptod{} is presented merely with two formatting sample turns from the source tasks/domains in the prompt.

\paragraph{Results.}
Following \citet{mehri2021schema, mosig2020star}, we report the zero-shot evaluation results in two settings, \ie task transfer and domain transfer in Table \ref{tab:star}. 
\textsc{SGP-TOD} demonstrates exceptional performance, surpassing or rivaling zero-shot transfer methods in both settings. This outcome signifies that, even when faced with complicated business logic and system actions in dialog policies, the proposed \textsc{SGP-TOD} continues to exhibit commendable performance.

\paragraph{\textit{Comparison with Zero-Shot Transfer Methods.}}
\textsc{SGP-TOD}, \textit{merely with two formatting sample turns}, achieves superior or comparable performance compared to \textsc{BERT+S}, \textsc{SAM}, which are fine-tuned on adequate source data. 
Figure \ref{zero-shot-star} shows the impact of changing the number of training dialogs (ranging from 1 to 1,000) and formatting example turns (spanning from 1 to 10) from source domains/tasks. We observe that \textsc{SGP-TOD}, \textit{employing only two formatting sample turns}, attains superior or commensurate performance in terms of both $\mathtt{F1 score}$ and $\mathtt{Accuracy}$ (as detailed in Appendix \ref{sec:figure_star_acc}), when compared to \textsc{SAM} trained with 1,000 dialogs. Given that a single dialog contains more than 10 dialogue turns, this result suggests that \textsc{SGP-TOD} diminishes labeling expenses by a minimum factor of 1,000. 
Furthermore, it is noteworthy that augmenting the quantity of formatting exemplar turns exerts a negligible influence on the performance of \textsc{SGP-TOD}.

\paragraph{\textit{Impact of Different LLMs and Prompting Formats.}}

\textsc{SGP-TOD-Codex} surpasses \textsc{SGP-TOD-GPT3.5} while rivaling \textsc{SGP-TOD-GPT3.5-E2E} (with template responses affixed to action labels in the policy prompt, demonstrated in Figure \ref{fig:e2e_policy_star} in Appendix \ref{sec:prompt_example}). We conjecture that Codex, benefiting from extensive pre-training on copious code data, demonstrates enhanced proficiency compared to GPT-3.5 in interpreting action labels. In addition, appending template responses is presumed to facilitate the explication of action labels for GPT-3.5.

\paragraph{\textit{Impact of Different Task Schema.}}

\textsc{SGP-TOD-Codex-INI}, utilizing an identical task schema as employed in training \textsc{SAM} \cite{mehri2021schema}, manifests commendable performance. This result highlights that \textsc{SGP-TOD} as a flexible prompting strategy, compatible with any manually-crafted task schema. Though the current study does not center around task schema design, future research endeavors may investigate the influence of varying task schemas, encompassing diverse formats and coverage.


\subsection{End-to-End Evaluation on Domain Extension}

\paragraph{Setup.}
We conduct experiments in a domain extension setting \cite{lipton2018bbq,DBLP:conf/interspeech/GasicKTBHSTY14} to assess the efficacy of \textsc{SGP-TOD} in adapting deployed task bots to incorporate novel functionalities. Following \citet{zhang2022toward}, we construct the \texttt{Restaurant-ext} corpus by extending the pre-existing \texttt{Restaurant} in \texttt{RADDLE} \cite{DBLP:conf/acl/PengLZZLG20} with additional functions. Specifically, we introduce four new slots: \textit{[restaurant\_dish]}, \textit{[value\_price]}, \textit{[start\_time]}, and \textit{[end\_time]}. The initial slot pertains to recommendations for signature restaurant meals, while the final three concern delivery service details. 
All database entries are updated with corresponding values. A dialog example and a DB entry in \texttt{Restaurant-ext} can be found in Appendix \ref{sec:task_extension}.
\begin{table}[!t]
\setlength\tabcolsep{3pt}
  \centering
  \begin{threeparttable}
  \fontsize{7.5}{8} 
  \selectfont
    \begin{tabular}{lccccc}
    \toprule
    \multirow{2}{*}{Model}&\multirow{2}{*}{FT/FS/ZS}&
    \multicolumn{4}{c}{\texttt{Restaurant-Ext}} \cr
    \cmidrule(lr){3-6}
    &&$\mathtt{Info.}$&$\mathtt{Succ.}$&$\mathtt{BLEU}$&$\mathtt{BERTS.}$ \cr
    \midrule
    \multicolumn{6}{l}{Without domain-relevant knowledge}\cr  
    \midrule
    ChatGPT &ZS& 44.00&6.00&4.31&85.96 \\
    GPT3.5 &ZS& 34.00&16.00&8.70&84.31 \\
    \midrule
    \multicolumn{6}{l}{With prior knowledge on \texttt{Restaurant}}\cr  
    \midrule
    \soloist{} &FT&78.00& 0.00&10.62 &87.24\\
    \textsc{SGP-TOD-ChatGPT} &ZS&88.00&34.00&5.45&86.11\\
    \textsc{SGP-TOD-GPT3.5} &ZS& 94.00&30.00&10.68& 87.30\\
    \midrule
    \multicolumn{6}{l}{With knowledge on \texttt{Restaurant-Ext}}\cr  
    \midrule
    \soloistteach &FT&82.00&38.00&10.99&87.66\\
    \textsc{Few-shot-GPT3.5} &FS&88.00&54.00&12.95&88.90\\
    \textsc{SGP-TOD-ChatGPT-Ext} &ZS&88.00&78.00&6.25&86.15\\
    \textsc{SGP-TOD-GPT3.5-Ext} &ZS&\textbf{96.00}&\textbf{86.00}&\textbf{14.57}&\textbf{89.01}\\

    \bottomrule  
    \end{tabular}
  \end{threeparttable}
  \caption{End-to-end evaluation on domain extension. FT: fine-tuning, FS: few-shot prompting, ZS: zero-shot prompting, Info.: Inform, Succ.: Success, BERTS.: BERTScore. (Difference in mean is significant with p<0.01.)}
  \label{tab:domain_extension}
  \vspace{-1mm}
\end{table}
\paragraph{Compared Methods.} 

\begin{itemize}\setlength{\itemsep}{0pt}
    \item ChatGPT, GPT-3.5 denote zero-shot prompting with base LLMs that receive merely two formatting example turns from other domains in \texttt{RADDLE}.\footnote{We utilize the same formatting example turns in all zero-shot prompting methods.}
    \item \textsc{SGP-TOD-ChatGPT}, \textsc{SGP-TOD-GPT3.5} represent our \textsc{SGP-TOD} implementation, with the \texttt{Restaurant} policy skeleton.
    \item \soloist{} is trained with 50 training dialogs in the \texttt{Restaurant} domain (previously reported in Table \ref{tab:results_4task}).
    \item \soloistteach{} is fine-tuning method enhanced with machine teaching \cite{DBLP:journals/corr/SimardACPGMRSVW17}. Machine teaching is an efficient approach to equip deployed task bots with the ability to handle new functions by correcting representative failed human-bot dialogs. 
    We deploy \soloist{} to converse with real users, then implement machine teaching via Conversational learner \cite{shukla-etal-2020-conversation}, an effective machine teaching tool, to obtain 10/50/50 examples in \texttt{Restaurant-ext} for training, validating, and testing. Finally, we fine-tune \soloist{} with gathered 10 training dialogs covering four new slots, resulting in dialog agent \soloistteach{}.
    \item \textsc{Few-shot-GPT3.5} is the few-shot prompting strategy augmented with machine teaching. Based on GPT-3.5, we utilize 10 randomly selected dialog turns from the collected 10 training dialogs as the prompt (with peak performance at 10), resulting in \textsc{Few-shot-GPT3.5}.
    \item \textsc{SGP-TOD-ChatGPT-Ext}, \textsc{SGP-TOD-GP3.5-Ext} refer to \textsc{SGP-TOD} with \texttt{Restaurant-Ext} policy skeleton, where we only add four template turns about four new slots to the policy skeleton of \texttt{Restaurant}.
\end{itemize}
 
\paragraph{Results.} The evaluation results are presented in Table \ref{tab:domain_extension}.  \textsc{SGP-TOD} with \texttt{Restaurant-Ext} policy skeleton, \ie \textsc{SGP-TOD-ChatGPT-Ext}, and notably \textsc{SGP-TOD-GPT3.5-Ext} surpasses all other evaluated approaches by a substantial margin over all the metrics. This demonstrates the adaptability of our \sptod{} in accommodating novel functionalities, thereby revealing its immense potential for lifelong learning.

\paragraph{\textit{Comparison with Approaches Augmented by Machine Teaching.}}
\textsc{SGP-TOD-GP3.5-Ext} obtains substantially higher $\mathtt{Success}$ rates than Few-shot-GPT3.5 (an increase of $32\%$) and \soloistteach{} (a rise of $48\%$). This remarkable zero-shot performance can be ascribed to the adequate coverage and exceptional interpretability of the task schema in \textsc{SGP-TOD}. Furthermore, in contrast to fine-tuning/prompting strategies utilizing additional dialogues corrected through machine teaching, \textsc{SGP-TOD} facilitates a more agile adaptation to novel functionalities by merely modifying template turns within the task schema.
\paragraph{\textit{Comparison with Fine-Tuning Methods.}}
\soloist{} yields zero $\mathtt{Success}$, a predictable result given its lack of awareness regarding the new features. Utilizing only prior knowledge of \texttt{Restaurant}, our \textsc{SGP-TOD} with the \texttt{Restaurant} policy skeleton, \ie textsc{SGP-TOD-ChatGPT} and \textsc{SGP-TOD-GP3.5} demonstrably surpasses \soloist{} in terms of $\mathtt{Inform}$ (by over 10 points) and $\mathtt{Success}$ (by more than 30 points). This illustrates that our \textsc{SGP-TOD} provides enhanced robustness in zero-shot generalization.

\paragraph{\textit{Comparison with Base LLMs.}}
The substantial improvement of \textsc{SGP-TOD-ChatGPT-Ext} and \textsc{SGP-TOD-GPT3.5-Ext} over ChatGPT and GPT-3.5 illustrates \sptod{}'s efficiency in supplying task-specific knowledge in a zero-shot way. 

\paragraph{\textit{Impact of Different LLMs.}}
\textsc{SGP-TOD-ChatGPT-Ext} attains a lower $\mathtt{BLEU}$ yet a comparable $\mathtt{BERTScore}$, suggesting that ChatGPT generates more diverse responses relative to GPT-3.5.


\section{Discussion}
\subsection{Ablation Study}

\begin{table*}[!t]
  \centering
  \begin{threeparttable}
  \fontsize{8}{9}
  \selectfont
    \begin{tabular}{lcccccccc}
    \toprule
    \multirow{2}{*}{Model}&
    \multicolumn{4}{c}{\texttt{Multiwoz 2.0}}
     &\multicolumn{4}{c}{\texttt{Multiwoz 2.2}}\cr\cmidrule(lr){2-5} \cmidrule(lr){6-9}
     &$\mathtt{Inform}$&$\mathtt{Success}$&$\mathtt{BLEU}$&$\mathtt{Combined}$&$\mathtt{Inform}$&$\mathtt{Success}$&$\mathtt{BLEU}$&$\mathtt{Combined}$ \cr
    \midrule
    SP-TOD-GPT3.5 &\textbf{83.88}&\textbf{69.87}&\textbf{9.09}&\textbf{85.97}&\textbf{82.00}&\textbf{72.50}&\textbf{9.22}&\textbf{86.47}\\
    \midrule
    -policy &82.28&55.65&6.51&75.48&81.80&56.20&6.63&75.63 \\
    -policy -DB&81.20&50.95&6.48&72.56&81.40&52.30&6.57&73.42\\
    -policy -DB -belief &38.74&33.13&6.18&42.12&38.60&33.90&6.29&42.54 \\
    \bottomrule  
    \end{tabular}
  \end{threeparttable}
  
  \caption{Ablation study on the impact of the three components in the proposed \sptod{} and the database expertise on \texttt{Multiwoz} using GPT-3.5. -policy: removing policy prompter, -DB: removing database retriever, -belief: removing DST prompter.}
  \label{tab:ablation}
  \vspace{-1mm}
\end{table*}

\begin{table*}[th]
\setlength\tabcolsep{1.5pt}
  \centering
  \begin{threeparttable}
  \fontsize{8.5}{9}
  \selectfont
    \begin{tabular}{l p{7cm} |l p{7cm}} 
    
    \toprule
    Turn  & \makecell[c]{Interaction between a user and ChatGPT} &
     Turn & \makecell[c]{Interaction between a user and SP-TOD-ChatGPT-Ext}\\
     
    \midrule
    1 & \makecell[l]{\textbf{User:} Hi, i'm looking for a place with \textit{Tuscan} food \\in \textit{any} price range.} &1& \makecell[l]{\textbf{User:} Hi, i'm looking for a place with \textit{Tuscan} food \\in \textit{any} price range.} \\
    2& \cellcolor{pink!70}{\makecell[l]{\textbf{System:} Certainly, there are \textcolor{red}{3 \textit{Tuscan} restaurants} \\in town. What is your \textcolor{red}{preferred price range}?}} & 2& \cellcolor{lime!60}{\makecell[l]{\textbf{System:} I am sorry, there are \textcolor{blue}{no \textit{Tuscan} restaurants}. \\ Here are \textcolor{blue}{some other options: \textit{Korean, Italian, Chinese}}. \\ Which one would you like more information on?}} \\
    3 & \makecell[l]{\textbf{User:} I \textit{don't care} about the price. Please recommend \\one for me, i also need the phone number please.} &3& \makecell[l]{\textbf{User:} How about any \textit{Korean} restaurants? \\I also need the phone number please.}  \\
    4& \cellcolor{pink!70}{\makecell[l]{\textbf{System:} The phone number for \textcolor{red}{\textit{Pizza Hut City Centre}} \\is \textcolor{red}{\textit{01223323737}}. Is there anything else i can help you \\with today?}} & 4& \cellcolor{lime!60}\makecell[l]{\textbf{System:} \textcolor{blue}{\textit{Little Seoul}} is a \textcolor{blue}{\textit{Korean}} restaurant. \\Their phone number is \textcolor{blue}{\textit{01223308681}}.} \\
    5 & \makecell[l]{\textbf{User:} Does the restaurant offer delivery service?\\ How much does the delivery charge?} &5& \makecell[l]{\textbf{User:} Does the restaurant offer delivery service?\\ How much does the delivery charge?} \\
    6& \cellcolor{pink!70}{\makecell[l]{\textbf{System:} I'm sorry, but \textcolor{red}{I could not find any information} \\\textcolor{red}{on delivery fee} in our database. You may want to contact \\ the restaurant directly to inquire about this. Is there \\anything else I can assist you with today?}} & 6& \cellcolor{lime!60}\makecell[l]{\textbf{System:} \textcolor{blue}{Yes}, they offer delivery service and the delivery \\charge is \textcolor{blue}{\textit{4 pounds}}. Can I help you with anything else?} \\
    7 & \makecell[l]{\textbf{User:} No. Thank you, goodbye.} &7& \makecell[l]{\textbf{User:} No. Thank you, goodbye.} \\
    8& \makecell[l]{\textbf{System:} Glad i could assist you. Have a good day.} & 8& \makecell[l]{\textbf{System:} You're welcome, have a great meal! Goodbye!} \\ 

    \bottomrule  
    \end{tabular}
  \end{threeparttable}
  
  \caption{Illustration of two interactive examples between a user and \textsc{ChatGPT} (left), \textsc{SGP-TOD-ChatGPT-Ext} (right). \colorbox{pink!70}{Appropriate responses} and \colorbox{lime!60}{Inappropriate responses} are highlighted.}
  \label{tab:case}
  \vspace{-1mm}
\end{table*}

In Table \ref{tab:ablation}, we study the impact of the three components of \sptod{} (namely, Policy Prompter, DST Prompter, and LLM) as well as the database expertise, on \texttt{Multiwoz} (2.0 and 2.2) utlizing GPT-3.5.\footnote{We inject the same two formatting example turns into the prompt throughout the evaluation.} Combining the three elements in \sptod{} with the database expertise produces optimal results across both datasets. The removal of the Policy Prompter, database knowledge, and DST Prompter leads to consistent declines in all evaluation metrics, underscoring the value of enhancing the fixed LLM with the task schema and external database information.


Specifically, \textbf{GPT-3.5} (in the final row) exhibits commendable zero-shot performance, highlighting the need of exploiting its superior zero-shot generalization capabilities in dialog generation tasks. Additionally, \textbf{Disabling the Policy Prompter} incurs a discernible decline in performance regarding $\mathtt{Success}$ (approximately $15\%$) and $\mathtt{BLEU}$ (roughly $3\%$), as the Policy Prompter's primary function is to provide task completion guidelines and interaction patterns. \textbf{Eliminating the database expertise} primarily reduces $\mathtt{Success}$ (by approximately $3\%$), implying that incorporating database information contributes to task completion. Lastly, \textbf{excising the DST Prompter} engenders a considerable diminution in performance concerning $\mathtt{Inform}$ (around $40\%$) and $\mathtt{Success}$ (nearly $17\%$), due to the DST Prompter's intended purpose of assisting the frozen LLM in apprehending the dialogue context.



\subsection{Case Study}

Despite the superior performance of the proposed \textsc{SGP-TOD} on GPT-3.5, we showcase interactive examples utilizing ChatGPT, a renowned and potent chatbot. In Table \ref{tab:case}, a user engages with ChatGPT (left) and \textsc{SGP-TOD-ChatGPT-Ext} (right) to complete the identical task on \texttt{Restaurant-Ext}.\footnote{ChatGPT and \textsc{SGP-TOD-ChatGPT-Ext} are previously reported in Table \ref{tab:domain_extension}. The same two formatting example turns are incorporated into the prompt for both zero-shot strategies.} The user initiates the conversation by seeking recommendations for a Tuscan restaurant with no price range preference. Lacking external database information, ChatGPT conveys inaccurate details (Turn 2), whereas \textsc{SGP-TOD-ChatGPT-Ext} informs users of the absence of matching restaurants and proposes alternatives (Turn 2). This exemplifies the benefits of integrating real-world expertise into the fixed LLM. Furthermore, ChatGPT persistently inquires about the desired price range despite the user's indifference. We argue that SGP-TOD assists the frozen LLM in discerning user intentions. In Turn 4, ChatGPT continues to furnish fabricated details (\ie the restaurant name and phone number) concerning the nonexistent eatery, while \textsc{SGP-TOD-ChatGPT-Ext} identifies a suitable Korean restaurant and the corresponding factual information. In contrast with ChatGPT, \textsc{SGP-TOD-ChatGPT-Ext} adeptly addresses inquiries about the delivery service (Turn 6), indicating that SGP-TOD is capable of endowing the frozen LLM with novel functionalities.

\section{Conclusion}
In this paper, we present \sptod{}, a schema-guided prompting strategy aimed at the expeditious construction of end-to-end task bots, relying exclusively on LLMs and the corresponding task schema.
Employing the symbolic knowledge -- the pre-defined task schema -- \sptod{} guides fixed LLMs to generate suitable responses for novel tasks in a zero-shot fashion. Experimental results on Multiwoz (2.0 and 2.2), RADDLE, and STAR in zero-shot settings show that \sptod{} attains SOTA zero-shot performance, substantially outpacing few-shot approaches and even even demonstrating competitive results in relation to full-shot fine-tuning methods. This suggests that our \sptod{} presents an appealing starting point for utilizing LLMs to construct task bots with minimal human intervention. In a domain-extension scenario, \sptod{} exhibits remarkable adaptability to new functionalities, showcasing its impressive potential for lifelong learning.
For future work, we plan to explore the use of \sptod{} to develop personalized chatbots by utilizing pertinent task schema.
\newpage
\bibliography{anthology,custom}

\begin{thebibliography}{65}
\expandafter\ifx\csname natexlab\endcsname\relax\def\natexlab#1{#1}\fi

\bibitem[{Brown et~al.(2020)Brown, Mann, Ryder, Subbiah, Kaplan, Dhariwal,
  Neelakantan, Shyam, Sastry, Askell, Agarwal, Herbert-Voss, Krueger, Henighan,
  Child, Ramesh, Ziegler, Wu, Winter, Hesse, Chen, Sigler, Litwin, Gray, Chess,
  Clark, Berner, McCandlish, Radford, Sutskever, and
  Amodei}]{NEURIPS2020_1457c0d6}
Tom Brown, Benjamin Mann, Nick Ryder, Melanie Subbiah, Jared~D Kaplan, Prafulla
  Dhariwal, Arvind Neelakantan, Pranav Shyam, Girish Sastry, Amanda Askell,
  Sandhini Agarwal, Ariel Herbert-Voss, Gretchen Krueger, Tom Henighan, Rewon
  Child, Aditya Ramesh, Daniel Ziegler, Jeffrey Wu, Clemens Winter, Chris
  Hesse, Mark Chen, Eric Sigler, Mateusz Litwin, Scott Gray, Benjamin Chess,
  Jack Clark, Christopher Berner, Sam McCandlish, Alec Radford, Ilya Sutskever,
  and Dario Amodei. 2020.
\newblock \href
  {https://proceedings.neurips.cc/paper/2020/file/1457c0d6bfcb4967418bfb8ac142f64a-Paper.pdf}
  {Language models are few-shot learners}.
\newblock In \emph{Advances in Neural Information Processing Systems},
  volume~33, pages 1877--1901. Curran Associates, Inc.

\bibitem[{Bubeck et~al.(2023)Bubeck, Chandrasekaran, Eldan, Gehrke, Horvitz,
  Kamar, Lee, Lee, Li, Lundberg, Nori, Palangi, Ribeiro, and
  Zhang}]{bubeck2023sparks}
Sébastien Bubeck, Varun Chandrasekaran, Ronen Eldan, Johannes Gehrke, Eric
  Horvitz, Ece Kamar, Peter Lee, Yin~Tat Lee, Yuanzhi Li, Scott Lundberg,
  Harsha Nori, Hamid Palangi, Marco~Tulio Ribeiro, and Yi~Zhang. 2023.
\newblock \href {http://arxiv.org/abs/2303.12712} {Sparks of artificial general
  intelligence: Early experiments with gpt-4}.

\bibitem[{Budzianowski et~al.(2018)Budzianowski, Wen, Tseng, Casanueva, Stefan,
  Osman, and Ga{\v{s}}i\'c}]{budzianowski2018large}
Pawe{\l} Budzianowski, Tsung-Hsien Wen, Bo-Hsiang Tseng, I{\~n}igo Casanueva,
  Ultes Stefan, Ramadan Osman, and Milica Ga{\v{s}}i\'c. 2018.
\newblock Multiwoz - a large-scale multi-domain wizard-of-oz dataset for
  task-oriented dialogue modelling.
\newblock In \emph{Proceedings of the 2018 Conference on Empirical Methods in
  Natural Language Processing (EMNLP)}.

\bibitem[{Campagna et~al.(2020)Campagna, Foryciarz, Moradshahi, and
  Lam}]{campagna-etal-2020-zero}
Giovanni Campagna, Agata Foryciarz, Mehrad Moradshahi, and Monica Lam. 2020.
\newblock \href {https://doi.org/10.18653/v1/2020.acl-main.12} {Zero-shot
  transfer learning with synthesized data for multi-domain dialogue state
  tracking}.
\newblock In \emph{Proceedings of the 58th Annual Meeting of the Association
  for Computational Linguistics}, pages 122--132, Online. Association for
  Computational Linguistics.

\bibitem[{Chen et~al.(2021)Chen, Tworek, Jun, Yuan, Pinto, Kaplan, Edwards,
  Burda, Joseph, Brockman et~al.}]{chen2021evaluating}
Mark Chen, Jerry Tworek, Heewoo Jun, Qiming Yuan, Henrique Ponde de~Oliveira
  Pinto, Jared Kaplan, Harri Edwards, Yuri Burda, Nicholas Joseph, Greg
  Brockman, et~al. 2021.
\newblock Evaluating large language models trained on code.
\newblock \emph{arXiv preprint arXiv:2107.03374}.

\bibitem[{Cheng et~al.(2023)Cheng, Xie, Shi, Li, Nadkarni, Hu, Xiong, Radev,
  Ostendorf, Zettlemoyer, Smith, and Yu}]{Binder}
Zhoujun Cheng, Tianbao Xie, Peng Shi, Chengzu Li, Rahul Nadkarni, Yushi Hu,
  Caiming Xiong, Dragomir Radev, Mari Ostendorf, Luke Zettlemoyer, Noah~A.
  Smith, and Tao Yu. 2023.
\newblock Binding language models in symbolic languages.
\newblock \emph{ICLR}, abs/2210.02875.

\bibitem[{Chowdhery et~al.(2022{\natexlab{a}})Chowdhery, Narang, Devlin, Bosma,
  Mishra, Roberts, Barham, Chung, Sutton, Gehrmann, Schuh, Shi, Tsvyashchenko,
  Maynez, Rao, Barnes, Tay, Shazeer, Prabhakaran, Reif, Du, Hutchinson, Pope,
  Bradbury, Austin, Isard, Gur-Ari, Yin, Duke, Levskaya, Ghemawat, Dev,
  Michalewski, Garcia, Misra, Robinson, Fedus, Zhou, Ippolito, Luan, Lim, Zoph,
  Spiridonov, Sepassi, Dohan, Agrawal, Omernick, Dai, Pillai, Pellat,
  Lewkowycz, Moreira, Child, Polozov, Lee, Zhou, Wang, Saeta, Diaz, Firat,
  Catasta, Wei, Meier-Hellstern, Eck, Dean, Petrov, and
  Fiedel}]{chowdhery2022palm}
Aakanksha Chowdhery, Sharan Narang, Jacob Devlin, Maarten Bosma, Gaurav Mishra,
  Adam Roberts, Paul Barham, Hyung~Won Chung, Charles Sutton, Sebastian
  Gehrmann, Parker Schuh, Kensen Shi, Sasha Tsvyashchenko, Joshua Maynez,
  Abhishek Rao, Parker Barnes, Yi~Tay, Noam Shazeer, Vinodkumar Prabhakaran,
  Emily Reif, Nan Du, Ben Hutchinson, Reiner Pope, James Bradbury, Jacob
  Austin, Michael Isard, Guy Gur-Ari, Pengcheng Yin, Toju Duke, Anselm
  Levskaya, Sanjay Ghemawat, Sunipa Dev, Henryk Michalewski, Xavier Garcia,
  Vedant Misra, Kevin Robinson, Liam Fedus, Denny Zhou, Daphne Ippolito, David
  Luan, Hyeontaek Lim, Barret Zoph, Alexander Spiridonov, Ryan Sepassi, David
  Dohan, Shivani Agrawal, Mark Omernick, Andrew~M. Dai,
  Thanumalayan~Sankaranarayana Pillai, Marie Pellat, Aitor Lewkowycz, Erica
  Moreira, Rewon Child, Oleksandr Polozov, Katherine Lee, Zongwei Zhou, Xuezhi
  Wang, Brennan Saeta, Mark Diaz, Orhan Firat, Michele Catasta, Jason Wei,
  Kathy Meier-Hellstern, Douglas Eck, Jeff Dean, Slav Petrov, and Noah Fiedel.
  2022{\natexlab{a}}.
\newblock \href {http://arxiv.org/abs/2204.02311} {Palm: Scaling language
  modeling with pathways}.

\bibitem[{Chowdhery et~al.(2022{\natexlab{b}})Chowdhery, Narang, Devlin, Bosma,
  Mishra, Roberts, Barham, Chung, Sutton, Gehrmann, Schuh, Shi, Tsvyashchenko,
  Maynez, Rao, Barnes, Tay, Shazeer, Prabhakaran, Reif, Du, Hutchinson, Pope,
  Bradbury, Austin, Isard, Gur{-}Ari, Yin, Duke, Levskaya, Ghemawat, Dev,
  Michalewski, Garcia, Misra, Robinson, Fedus, Zhou, Ippolito, Luan, Lim, Zoph,
  Spiridonov, Sepassi, Dohan, Agrawal, Omernick, Dai, Pillai, Pellat,
  Lewkowycz, Moreira, Child, Polozov, Lee, Zhou, Wang, Saeta, Diaz, Firat,
  Catasta, Wei, Meier{-}Hellstern, Eck, Dean, Petrov, and
  Fiedel}]{DBLP:journals/corr/abs-2204-02311}
Aakanksha Chowdhery, Sharan Narang, Jacob Devlin, Maarten Bosma, Gaurav Mishra,
  Adam Roberts, Paul Barham, Hyung~Won Chung, Charles Sutton, Sebastian
  Gehrmann, Parker Schuh, Kensen Shi, Sasha Tsvyashchenko, Joshua Maynez,
  Abhishek Rao, Parker Barnes, Yi~Tay, Noam Shazeer, Vinodkumar Prabhakaran,
  Emily Reif, Nan Du, Ben Hutchinson, Reiner Pope, James Bradbury, Jacob
  Austin, Michael Isard, Guy Gur{-}Ari, Pengcheng Yin, Toju Duke, Anselm
  Levskaya, Sanjay Ghemawat, Sunipa Dev, Henryk Michalewski, Xavier Garcia,
  Vedant Misra, Kevin Robinson, Liam Fedus, Denny Zhou, Daphne Ippolito, David
  Luan, Hyeontaek Lim, Barret Zoph, Alexander Spiridonov, Ryan Sepassi, David
  Dohan, Shivani Agrawal, Mark Omernick, Andrew~M. Dai,
  Thanumalayan~Sankaranarayana Pillai, Marie Pellat, Aitor Lewkowycz, Erica
  Moreira, Rewon Child, Oleksandr Polozov, Katherine Lee, Zongwei Zhou, Xuezhi
  Wang, Brennan Saeta, Mark Diaz, Orhan Firat, Michele Catasta, Jason Wei,
  Kathy Meier{-}Hellstern, Douglas Eck, Jeff Dean, Slav Petrov, and Noah
  Fiedel. 2022{\natexlab{b}}.
\newblock \href {https://doi.org/10.48550/arXiv.2204.02311} {Palm: Scaling
  language modeling with pathways}.
\newblock \emph{CoRR}, abs/2204.02311.

\bibitem[{Dai et~al.(2023)Dai, Liu, Liao, Huang, Cao, Wu, Zhao, Xu, Liu, Liu,
  Li, Zhu, Cai, Sun, Li, Shen, Liu, and Li}]{dai2023auggpt}
Haixing Dai, Zhengliang Liu, Wenxiong Liao, Xiaoke Huang, Yihan Cao, Zihao Wu,
  Lin Zhao, Shaochen Xu, Wei Liu, Ninghao Liu, Sheng Li, Dajiang Zhu, Hongmin
  Cai, Lichao Sun, Quanzheng Li, Dinggang Shen, Tianming Liu, and Xiang Li.
  2023.
\newblock \href {http://arxiv.org/abs/2302.13007} {Auggpt: Leveraging chatgpt
  for text data augmentation}.

\bibitem[{Devlin et~al.(2019)Devlin, Chang, Lee, and
  Toutanova}]{DBLP:conf/naacl/DevlinCLT19}
Jacob Devlin, Ming{-}Wei Chang, Kenton Lee, and Kristina Toutanova. 2019.
\newblock \href {https://doi.org/10.18653/v1/n19-1423} {{BERT:} pre-training of
  deep bidirectional transformers for language understanding}.
\newblock In \emph{Proceedings of the 2019 Conference of the North American
  Chapter of the Association for Computational Linguistics: Human Language
  Technologies, {NAACL-HLT} 2019, Minneapolis, MN, USA, June 2-7, 2019, Volume
  1 (Long and Short Papers)}, pages 4171--4186. Association for Computational
  Linguistics.

\bibitem[{Dong et~al.(2023)Dong, Li, Dai, Zheng, Wu, Chang, Sun, Xu, Li, and
  Sui}]{dong2023survey}
Qingxiu Dong, Lei Li, Damai Dai, Ce~Zheng, Zhiyong Wu, Baobao Chang, Xu~Sun,
  Jingjing Xu, Lei Li, and Zhifang Sui. 2023.
\newblock \href {http://arxiv.org/abs/2301.00234} {A survey on in-context
  learning}.

\bibitem[{Gasic et~al.(2014)Gasic, Kim, Tsiakoulis, Breslin, Henderson,
  Szummer, Thomson, and Young}]{DBLP:conf/interspeech/GasicKTBHSTY14}
Milica Gasic, Dongho Kim, Pirros Tsiakoulis, Catherine Breslin, Matthew
  Henderson, Martin Szummer, Blaise Thomson, and Steve~J. Young. 2014.
\newblock \href
  {http://www.isca-speech.org/archive/interspeech\_2014/i14\_0140.html}
  {Incremental on-line adaptation of pomdp-based dialogue managers to extended
  domains}.
\newblock In \emph{{INTERSPEECH} 2014, 15th Annual Conference of the
  International Speech Communication Association, Singapore, September 14-18,
  2014}, pages 140--144. {ISCA}.

\bibitem[{He et~al.(2022)He, Dai, Zheng, Wu, Cao, Liu, Jiang, Yang, Huang, Si
  et~al.}]{he2022galaxy}
Wanwei He, Yinpei Dai, Yinhe Zheng, Yuchuan Wu, Zheng Cao, Dermot Liu, Peng
  Jiang, Min Yang, Fei Huang, Luo Si, et~al. 2022.
\newblock Galaxy: A generative pre-trained model for task-oriented dialog with
  semi-supervised learning and explicit policy injection.
\newblock \emph{Proceedings of the AAAI Conference on Artificial Intelligence}.

\bibitem[{Hosseini-Asl et~al.(2020)Hosseini-Asl, McCann, Wu, Yavuz, and
  Socher}]{hosseini2020simple}
Ehsan Hosseini-Asl, Bryan McCann, Chien-Sheng Wu, Semih Yavuz, and Richard
  Socher. 2020.
\newblock A simple language model for task-oriented dialogue.
\newblock \emph{Advances in Neural Information Processing Systems},
  33:20179--20191.

\bibitem[{Hu et~al.(2022)Hu, Lee, Xie, Yu, Smith, and
  Ostendorf}]{hu-etal-2022-context}
Yushi Hu, Chia-Hsuan Lee, Tianbao Xie, Tao Yu, Noah~A. Smith, and Mari
  Ostendorf. 2022.
\newblock \href {https://aclanthology.org/2022.findings-emnlp.193} {In-context
  learning for few-shot dialogue state tracking}.
\newblock In \emph{Findings of the Association for Computational Linguistics:
  EMNLP 2022}, pages 2627--2643, Abu Dhabi, United Arab Emirates. Association
  for Computational Linguistics.

\bibitem[{Hudecek and Dusek(2023)}]{DBLP:journals/corr/abs-2304-06556}
Vojtech Hudecek and Ondrej Dusek. 2023.
\newblock \href {https://doi.org/10.48550/arXiv.2304.06556} {Are llms all you
  need for task-oriented dialogue?}
\newblock \emph{CoRR}, abs/2304.06556.

\bibitem[{Kale and Rastogi(2020)}]{DBLP:conf/emnlp/KaleR20}
Mihir Kale and Abhinav Rastogi. 2020.
\newblock \href {https://doi.org/10.18653/v1/2020.emnlp-main.527} {Template
  guided text generation for task-oriented dialogue}.
\newblock In \emph{Proceedings of the 2020 Conference on Empirical Methods in
  Natural Language Processing, {EMNLP} 2020, Online, November 16-20, 2020},
  pages 6505--6520. Association for Computational Linguistics.

\bibitem[{Li et~al.(2022)Li, Chen, Li, Wang, Qian, and
  Yan}]{li-etal-2022-controllable}
Zekun Li, Wenhu Chen, Shiyang Li, Hong Wang, Jing Qian, and Xifeng Yan. 2022.
\newblock \href {https://aclanthology.org/2022.findings-emnlp.318}
  {Controllable dialogue simulation with in-context learning}.
\newblock In \emph{Findings of the Association for Computational Linguistics:
  EMNLP 2022}, pages 4330--4347, Abu Dhabi, United Arab Emirates. Association
  for Computational Linguistics.

\bibitem[{Liang et~al.(2023)Liang, Wu, Song, Wu, Xia, Liu, Ou, Lu, Ji, Mao,
  Wang, Shou, Gong, and Duan}]{liang2023taskmatrixai}
Yaobo Liang, Chenfei Wu, Ting Song, Wenshan Wu, Yan Xia, Yu~Liu, Yang Ou, Shuai
  Lu, Lei Ji, Shaoguang Mao, Yun Wang, Linjun Shou, Ming Gong, and Nan Duan.
  2023.
\newblock \href {http://arxiv.org/abs/2303.16434} {Taskmatrix.ai: Completing
  tasks by connecting foundation models with millions of apis}.

\bibitem[{Lin et~al.(2021{\natexlab{a}})Lin, Liu, Madotto, Moon, Zhou, Crook,
  Wang, Yu, Cho, Subba et~al.}]{lin2021zero}
Zhaojiang Lin, Bing Liu, Andrea Madotto, Seungwhan Moon, Zhenpeng Zhou, Paul~A
  Crook, Zhiguang Wang, Zhou Yu, Eunjoon Cho, Rajen Subba, et~al.
  2021{\natexlab{a}}.
\newblock Zero-shot dialogue state tracking via cross-task transfer.
\newblock In \emph{Proceedings of the 2021 Conference on Empirical Methods in
  Natural Language Processing}, pages 7890--7900.

\bibitem[{Lin et~al.(2021{\natexlab{b}})Lin, Liu, Moon, Crook, Zhou, Wang, Yu,
  Madotto, Cho, and Subba}]{lin2021leveraging}
Zhaojiang Lin, Bing Liu, Seungwhan Moon, Paul~A Crook, Zhenpeng Zhou, Zhiguang
  Wang, Zhou Yu, Andrea Madotto, Eunjoon Cho, and Rajen Subba.
  2021{\natexlab{b}}.
\newblock Leveraging slot descriptions for zero-shot cross-domain dialogue
  statetracking.
\newblock In \emph{Proceedings of the 2021 Conference of the North American
  Chapter of the Association for Computational Linguistics: Human Language
  Technologies}, pages 5640--5648.

\bibitem[{Lin et~al.(2020)Lin, Madotto, Winata, and Fung}]{lin-etal-2020-mintl}
Zhaojiang Lin, Andrea Madotto, Genta~Indra Winata, and Pascale Fung. 2020.
\newblock \href {https://doi.org/10.18653/v1/2020.emnlp-main.273} {{M}in{TL}:
  Minimalist transfer learning for task-oriented dialogue systems}.
\newblock In \emph{Proceedings of the 2020 Conference on Empirical Methods in
  Natural Language Processing (EMNLP)}, pages 3391--3405, Online. Association
  for Computational Linguistics.

\bibitem[{Lipton et~al.(2018)Lipton, Li, Gao, Li, Ahmed, and
  Deng}]{lipton2018bbq}
Zachary Lipton, Xiujun Li, Jianfeng Gao, Lihong Li, Faisal Ahmed, and Li~Deng.
  2018.
\newblock Bbq-networks: Efficient exploration in deep reinforcement learning
  for task-oriented dialogue systems.
\newblock In \emph{Proceedings of the AAAI Conference on Artificial
  Intelligence}, volume~32.

\bibitem[{Liu et~al.(2022)Liu, Shen, Zhang, Dolan, Carin, and
  Chen}]{liu-etal-2022-makes}
Jiachang Liu, Dinghan Shen, Yizhe Zhang, Bill Dolan, Lawrence Carin, and Weizhu
  Chen. 2022.
\newblock \href {https://doi.org/10.18653/v1/2022.deelio-1.10} {What makes good
  in-context examples for {GPT}-3?}
\newblock In \emph{Proceedings of Deep Learning Inside Out (DeeLIO 2022): The
  3rd Workshop on Knowledge Extraction and Integration for Deep Learning
  Architectures}, pages 100--114, Dublin, Ireland and Online. Association for
  Computational Linguistics.

\bibitem[{Madotto et~al.(2021)Madotto, Lin, Winata, and Fung}]{madotto2021few}
Andrea Madotto, Zhaojiang Lin, Genta~Indra Winata, and Pascale Fung. 2021.
\newblock Few-shot bot: Prompt-based learning for dialogue systems.
\newblock \emph{arXiv preprint arXiv:2110.08118}.

\bibitem[{Mehri et~al.(2022)Mehri, Altun, and Eskenazi}]{mehri-etal-2022-lad}
Shikib Mehri, Yasemin Altun, and Maxine Eskenazi. 2022.
\newblock \href {https://aclanthology.org/2022.sigdial-1.55} {{LAD}: Language
  models as data for zero-shot dialog}.
\newblock In \emph{Proceedings of the 23rd Annual Meeting of the Special
  Interest Group on Discourse and Dialogue}, pages 595--604, Edinburgh, UK.
  Association for Computational Linguistics.

\bibitem[{Mehri and Eskenazi(2021)}]{mehri2021schema}
Shikib Mehri and Maxine Eskenazi. 2021.
\newblock Schema-guided paradigm for zero-shot dialog.
\newblock \emph{arXiv preprint arXiv:2106.07056}.

\bibitem[{Mosig et~al.(2020)Mosig, Mehri, and Kober}]{mosig2020star}
Johannes~EM Mosig, Shikib Mehri, and Thomas Kober. 2020.
\newblock Star: A schema-guided dialog dataset for transfer learning.
\newblock \emph{arXiv preprint arXiv:2010.11853}.

\bibitem[{Nye et~al.(2021)Nye, Tessler, Tenenbaum, and
  Lake}]{DBLP:conf/nips/NyeTTL21}
Maxwell~I. Nye, Michael~Henry Tessler, Joshua~B. Tenenbaum, and Brenden~M.
  Lake. 2021.
\newblock \href
  {https://proceedings.neurips.cc/paper/2021/hash/d3e2e8f631bd9336ed25b8162aef8782-Abstract.html}
  {Improving coherence and consistency in neural sequence models with
  dual-system, neuro-symbolic reasoning}.
\newblock In \emph{Advances in Neural Information Processing Systems 34: Annual
  Conference on Neural Information Processing Systems 2021, NeurIPS 2021,
  December 6-14, 2021, virtual}, pages 25192--25204.

\bibitem[{OpenAI(2023)}]{openai2023gpt4}
OpenAI. 2023.
\newblock \href {http://arxiv.org/abs/2303.08774} {Gpt-4 technical report}.

\bibitem[{Ouyang et~al.(2022)Ouyang, Wu, Jiang, Almeida, Wainwright, Mishkin,
  Zhang, Agarwal, Slama, Ray, Schulman, Hilton, Kelton, Miller, Simens, Askell,
  Welinder, Christiano, Leike, and Lowe}]{instructgpt2022}
Long Ouyang, Jeffrey Wu, Xu~Jiang, Diogo Almeida, Carroll Wainwright, Pamela
  Mishkin, Chong Zhang, Sandhini Agarwal, Katarina Slama, Alex Ray, John
  Schulman, Jacob Hilton, Fraser Kelton, Luke Miller, Maddie Simens, Amanda
  Askell, Peter Welinder, Paul~F Christiano, Jan Leike, and Ryan Lowe. 2022.
\newblock \href
  {https://proceedings.neurips.cc/paper_files/paper/2022/file/b1efde53be364a73914f58805a001731-Paper-Conference.pdf}
  {Training language models to follow instructions with human feedback}.
\newblock In \emph{Advances in Neural Information Processing Systems},
  volume~35, pages 27730--27744. Curran Associates, Inc.

\bibitem[{Papineni et~al.(2002)Papineni, Roukos, Ward, and
  Zhu}]{papineni-etal-2002-bleu}
Kishore Papineni, Salim Roukos, Todd Ward, and Wei-Jing Zhu. 2002.
\newblock \href {https://doi.org/10.3115/1073083.1073135} {{B}leu: a method for
  automatic evaluation of machine translation}.
\newblock In \emph{Proceedings of the 40th Annual Meeting of the Association
  for Computational Linguistics}, pages 311--318, Philadelphia, Pennsylvania,
  USA. Association for Computational Linguistics.

\bibitem[{Peng et~al.(2023)Peng, Galley, He, Cheng, Xie, Hu, Huang, Liden, Yu,
  Chen et~al.}]{peng2023check}
Baolin Peng, Michel Galley, Pengcheng He, Hao Cheng, Yujia Xie, Yu~Hu, Qiuyuan
  Huang, Lars Liden, Zhou Yu, Weizhu Chen, et~al. 2023.
\newblock Check your facts and try again: Improving large language models with
  external knowledge and automated feedback.
\newblock \emph{arXiv preprint arXiv:2302.12813}.

\bibitem[{Peng et~al.(2021{\natexlab{a}})Peng, Li, Li, Shayandeh, Liden, and
  Gao}]{peng2021soloist}
Baolin Peng, Chunyuan Li, Jinchao Li, Shahin Shayandeh, Lars Liden, and
  Jianfeng Gao. 2021{\natexlab{a}}.
\newblock Soloist: Building task bots at scale with transfer learning and
  machine teaching.
\newblock \emph{Transactions of the Association for Computational Linguistics},
  9:807--824.

\bibitem[{Peng et~al.(2021{\natexlab{b}})Peng, Li, Zhang, Li, Zhu, and
  Gao}]{peng2021synergy}
Baolin Peng, Chunyuan Li, Zhu Zhang, Jinchao Li, Chenguang Zhu, and Jianfeng
  Gao. 2021{\natexlab{b}}.
\newblock Synergy: Building task bots at scale using symbolic knowledge and
  machine teaching.
\newblock \emph{arXiv preprint arXiv:2110.11514}.

\bibitem[{Peng et~al.(2021{\natexlab{c}})Peng, Li, Zhang, Zhu, Li, and
  Gao}]{DBLP:conf/acl/PengLZZLG20}
Baolin Peng, Chunyuan Li, Zhu Zhang, Chenguang Zhu, Jinchao Li, and Jianfeng
  Gao. 2021{\natexlab{c}}.
\newblock \href {https://doi.org/10.18653/v1/2021.acl-long.341} {{RADDLE:} an
  evaluation benchmark and analysis platform for robust task-oriented dialog
  systems}.
\newblock In \emph{Proceedings of the 59th Annual Meeting of the Association
  for Computational Linguistics and the 11th International Joint Conference on
  Natural Language Processing, {ACL/IJCNLP} 2021, (Volume 1: Long Papers),
  Virtual Event, August 1-6, 2021}, pages 4418--4429. Association for
  Computational Linguistics.

\bibitem[{Qian and Yu(2019)}]{qian-yu-2019-domain}
Kun Qian and Zhou Yu. 2019.
\newblock \href {https://doi.org/10.18653/v1/P19-1253} {Domain adaptive dialog
  generation via meta learning}.
\newblock In \emph{Proceedings of the 57th Annual Meeting of the Association
  for Computational Linguistics}, pages 2639--2649, Florence, Italy.
  Association for Computational Linguistics.

\bibitem[{Qin et~al.(2023)Qin, Zhang, Zhang, Chen, Yasunaga, and
  Yang}]{qin2023chatgpt}
Chengwei Qin, Aston Zhang, Zhuosheng Zhang, Jiaao Chen, Michihiro Yasunaga, and
  Diyi Yang. 2023.
\newblock \href {http://arxiv.org/abs/2302.06476} {Is chatgpt a general-purpose
  natural language processing task solver?}

\bibitem[{Radford et~al.(2019)Radford, Wu, Child, Luan, Amodei, Sutskever
  et~al.}]{radford2019language}
Alec Radford, Jeffrey Wu, Rewon Child, David Luan, Dario Amodei, Ilya
  Sutskever, et~al. 2019.
\newblock Language models are unsupervised multitask learners.
\newblock \emph{OpenAI blog}, 1(8):9.

\bibitem[{Raffel et~al.(2020)Raffel, Shazeer, Roberts, Lee, Narang, Matena,
  Zhou, Li, and Liu}]{2020t5}
Colin Raffel, Noam Shazeer, Adam Roberts, Katherine Lee, Sharan Narang, Michael
  Matena, Yanqi Zhou, Wei Li, and Peter~J. Liu. 2020.
\newblock \href {http://jmlr.org/papers/v21/20-074.html} {Exploring the limits
  of transfer learning with a unified text-to-text transformer}.
\newblock \emph{Journal of Machine Learning Research}, 21(140):1--67.

\bibitem[{Rastogi et~al.(2020{\natexlab{a}})Rastogi, Zang, Sunkara, Gupta, and
  Khaitan}]{rastogi2020towards}
Abhinav Rastogi, Xiaoxue Zang, Srinivas Sunkara, Raghav Gupta, and Pranav
  Khaitan. 2020{\natexlab{a}}.
\newblock Towards scalable multi-domain conversational agents: The
  schema-guided dialogue dataset.
\newblock In \emph{Proceedings of the AAAI Conference on Artificial
  Intelligence}, volume~34, pages 8689--8696.

\bibitem[{Rastogi et~al.(2020{\natexlab{b}})Rastogi, Zang, Sunkara, Gupta, and
  Khaitan}]{DBLP:conf/aaai/RastogiZSGK20}
Abhinav Rastogi, Xiaoxue Zang, Srinivas Sunkara, Raghav Gupta, and Pranav
  Khaitan. 2020{\natexlab{b}}.
\newblock \href {https://ojs.aaai.org/index.php/AAAI/article/view/6394}
  {Towards scalable multi-domain conversational agents: The schema-guided
  dialogue dataset}.
\newblock In \emph{The Thirty-Fourth {AAAI} Conference on Artificial
  Intelligence, {AAAI} 2020, The Thirty-Second Innovative Applications of
  Artificial Intelligence Conference, {IAAI} 2020, The Tenth {AAAI} Symposium
  on Educational Advances in Artificial Intelligence, {EAAI} 2020, New York,
  NY, USA, February 7-12, 2020}, pages 8689--8696. {AAAI} Press.

\bibitem[{Roberts et~al.(2022)Roberts, Chung, Levskaya, Mishra, Bradbury,
  Andor, Narang, Lester, Gaffney, Mohiuddin, Hawthorne, Lewkowycz, Salcianu,
  van Zee, Austin, Goodman, Soares, Hu, Tsvyashchenko, Chowdhery, Bastings,
  Bulian, Garcia, Ni, Chen, Kenealy, Clark, Lee, Garrette, Lee-Thorp, Raffel,
  Shazeer, Ritter, Bosma, Passos, Maitin-Shepard, Fiedel, Omernick, Saeta,
  Sepassi, Spiridonov, Newlan, and Gesmundo}]{roberts2022scaling}
Adam Roberts, Hyung~Won Chung, Anselm Levskaya, Gaurav Mishra, James Bradbury,
  Daniel Andor, Sharan Narang, Brian Lester, Colin Gaffney, Afroz Mohiuddin,
  Curtis Hawthorne, Aitor Lewkowycz, Alex Salcianu, Marc van Zee, Jacob Austin,
  Sebastian Goodman, Livio~Baldini Soares, Haitang Hu, Sasha Tsvyashchenko,
  Aakanksha Chowdhery, Jasmijn Bastings, Jannis Bulian, Xavier Garcia, Jianmo
  Ni, Andrew Chen, Kathleen Kenealy, Jonathan~H. Clark, Stephan Lee, Dan
  Garrette, James Lee-Thorp, Colin Raffel, Noam Shazeer, Marvin Ritter, Maarten
  Bosma, Alexandre Passos, Jeremy Maitin-Shepard, Noah Fiedel, Mark Omernick,
  Brennan Saeta, Ryan Sepassi, Alexander Spiridonov, Joshua Newlan, and Andrea
  Gesmundo. 2022.
\newblock \href {http://arxiv.org/abs/2203.17189} {Scaling up models and data
  with $\texttt{t5x}$ and $\texttt{seqio}$}.

\bibitem[{Shah et~al.(2019)Shah, Gupta, Fayazi, and
  Hakkani-Tur}]{shah2019robust}
Darsh~J Shah, Raghav Gupta, Amir~A Fayazi, and Dilek Hakkani-Tur. 2019.
\newblock Robust zero-shot cross-domain slot filling with example values.
\newblock \emph{arXiv preprint arXiv:1906.06870}.

\bibitem[{Shin and Van~Durme(2022)}]{shin-van-durme-2022-shot}
Richard Shin and Benjamin Van~Durme. 2022.
\newblock \href {https://doi.org/10.18653/v1/2022.naacl-main.396} {Few-shot
  semantic parsing with language models trained on code}.
\newblock In \emph{Proceedings of the 2022 Conference of the North American
  Chapter of the Association for Computational Linguistics: Human Language
  Technologies}, pages 5417--5425, Seattle, United States. Association for
  Computational Linguistics.

\bibitem[{Shukla et~al.(2020)Shukla, Liden, Shayandeh, Kamal, Li, Mazzola,
  Park, Peng, and Gao}]{shukla-etal-2020-conversation}
Swadheen Shukla, Lars Liden, Shahin Shayandeh, Eslam Kamal, Jinchao Li, Matt
  Mazzola, Thomas Park, Baolin Peng, and Jianfeng Gao. 2020.
\newblock \href {https://doi.org/10.18653/v1/2020.acl-demos.39} {{C}onversation
  {L}earner - a machine teaching tool for building dialog managers for
  task-oriented dialog systems}.
\newblock In \emph{Proceedings of the 58th Annual Meeting of the Association
  for Computational Linguistics: System Demonstrations}, pages 343--349,
  Online. Association for Computational Linguistics.

\bibitem[{Simard et~al.(2017)Simard, Amershi, Chickering, Pelton, Ghorashi,
  Meek, Ramos, Suh, Verwey, Wang, and
  Wernsing}]{DBLP:journals/corr/SimardACPGMRSVW17}
Patrice~Y. Simard, Saleema Amershi, David~Maxwell Chickering, Alicia~Edelman
  Pelton, Soroush Ghorashi, Christopher Meek, Gonzalo~A. Ramos, Jina Suh, Johan
  Verwey, Mo~Wang, and John Wernsing. 2017.
\newblock \href {http://arxiv.org/abs/1707.06742} {Machine teaching: {A} new
  paradigm for building machine learning systems}.
\newblock \emph{CoRR}, abs/1707.06742.

\bibitem[{Su et~al.(2022)Su, Shu, Mansimov, Gupta, Cai, Lai, and
  Zhang}]{su2021multitask}
Yixuan Su, Lei Shu, Elman Mansimov, Arshit Gupta, Deng Cai, Yi{-}An Lai, and
  Yi~Zhang. 2022.
\newblock \href {https://arxiv.org/abs/2109.14739} {Multi-task pre-training for
  plug-and-play task-oriented dialogue system}.
\newblock \emph{Proceedings of the 60th Annual Meeting of the Association for
  Computational Linguistics (ACL)}.

\bibitem[{Sun et~al.(2022)Sun, Bao, Wu, and He}]{sun2022mars}
Haipeng Sun, Junwei Bao, Youzheng Wu, and Xiaodong He. 2022.
\newblock Mars: Semantic-aware contrastive learning for end-to-end
  task-oriented dialog.
\newblock \emph{arXiv preprint arXiv:2210.08917}.

\bibitem[{Wang et~al.(2023)Wang, Hu, Hou, Chen, Zheng, Wang, Yang, Huang, Ye,
  Geng, Jiao, Zhang, and Xie}]{wang2023robustness}
Jindong Wang, Xixu Hu, Wenxin Hou, Hao Chen, Runkai Zheng, Yidong Wang, Linyi
  Yang, Haojun Huang, Wei Ye, Xiubo Geng, Binxin Jiao, Yue Zhang, and Xing Xie.
  2023.
\newblock \href {http://arxiv.org/abs/2302.12095} {On the robustness of
  chatgpt: An adversarial and out-of-distribution perspective}.

\bibitem[{Wang et~al.(2022{\natexlab{a}})Wang, Cao, Li, Fu, Lin, and
  Guo}]{wang-etal-2022-slot}
Qingyue Wang, Yanan Cao, Piji Li, Yanhe Fu, Zheng Lin, and Li~Guo.
  2022{\natexlab{a}}.
\newblock \href {https://aclanthology.org/2022.coling-1.42} {Slot dependency
  modeling for zero-shot cross-domain dialogue state tracking}.
\newblock In \emph{Proceedings of the 29th International Conference on
  Computational Linguistics}, pages 510--520, Gyeongju, Republic of Korea.
  International Committee on Computational Linguistics.

\bibitem[{Wang et~al.(2022{\natexlab{b}})Wang, Mishra, Alipoormolabashi, Kordi,
  Mirzaei, Naik, Ashok, Dhanasekaran, Arunkumar, Stap et~al.}]{wang2022super}
Yizhong Wang, Swaroop Mishra, Pegah Alipoormolabashi, Yeganeh Kordi, Amirreza
  Mirzaei, Atharva Naik, Arjun Ashok, Arut~Selvan Dhanasekaran, Anjana
  Arunkumar, David Stap, et~al. 2022{\natexlab{b}}.
\newblock Super-naturalinstructions: Generalization via declarative
  instructions on 1600+ nlp tasks.
\newblock In \emph{Proceedings of the 2022 Conference on Empirical Methods in
  Natural Language Processing}, pages 5085--5109.

\bibitem[{Wei et~al.(2022)Wei, Tay, Bommasani, Raffel, Zoph, Borgeaud,
  Yogatama, Bosma, Zhou, Metzler, Chi, Hashimoto, Vinyals, Liang, Dean, and
  Fedus}]{wei2022emergent}
Jason Wei, Yi~Tay, Rishi Bommasani, Colin Raffel, Barret Zoph, Sebastian
  Borgeaud, Dani Yogatama, Maarten Bosma, Denny Zhou, Donald Metzler, Ed~H.
  Chi, Tatsunori Hashimoto, Oriol Vinyals, Percy Liang, Jeff Dean, and William
  Fedus. 2022.
\newblock \href {http://arxiv.org/abs/2206.07682} {Emergent abilities of large
  language models}.

\bibitem[{Williams and Liden(2017)}]{DBLP:conf/sigdial/WilliamsL17}
Jason~D. Williams and Lars Liden. 2017.
\newblock \href {https://doi.org/10.18653/v1/w17-5511} {Demonstration of
  interactive teaching for end-to-end dialog control with hybrid code
  networks}.
\newblock In \emph{Proceedings of the 18th Annual SIGdial Meeting on Discourse
  and Dialogue, Saarbr{\"{u}}cken, Germany, August 15-17, 2017}, pages 82--85.
  Association for Computational Linguistics.

\bibitem[{Wu et~al.(2019)Wu, Madotto, Hosseini-Asl, Xiong, Socher, and
  Fung}]{wu-etal-2019-transferable}
Chien-Sheng Wu, Andrea Madotto, Ehsan Hosseini-Asl, Caiming Xiong, Richard
  Socher, and Pascale Fung. 2019.
\newblock \href {https://doi.org/10.18653/v1/P19-1078} {Transferable
  multi-domain state generator for task-oriented dialogue systems}.
\newblock In \emph{Proceedings of the 57th Annual Meeting of the Association
  for Computational Linguistics}, pages 808--819, Florence, Italy. Association
  for Computational Linguistics.

\bibitem[{Xie et~al.(2022)Xie, Wu, Shi, Zhong, Scholak, Yasunaga, Wu, Zhong,
  Yin, Wang, Zhong, Wang, Li, Boyle, Ni, Yao, Radev, Xiong, Kong, Zhang, Smith,
  Zettlemoyer, and Yu}]{xie-etal-2022-unifiedskg}
Tianbao Xie, Chen~Henry Wu, Peng Shi, Ruiqi Zhong, Torsten Scholak, Michihiro
  Yasunaga, Chien-Sheng Wu, Ming Zhong, Pengcheng Yin, Sida~I. Wang, Victor
  Zhong, Bailin Wang, Chengzu Li, Connor Boyle, Ansong Ni, Ziyu Yao, Dragomir
  Radev, Caiming Xiong, Lingpeng Kong, Rui Zhang, Noah~A. Smith, Luke
  Zettlemoyer, and Tao Yu. 2022.
\newblock \href {https://aclanthology.org/2022.emnlp-main.39} {{U}nified{SKG}:
  Unifying and multi-tasking structured knowledge grounding with text-to-text
  language models}.
\newblock In \emph{Proceedings of the 2022 Conference on Empirical Methods in
  Natural Language Processing}, pages 602--631, Abu Dhabi, United Arab
  Emirates. Association for Computational Linguistics.

\bibitem[{Yang et~al.(2021)Yang, Li, and Quan}]{yang2021ubar}
Yunyi Yang, Yunhao Li, and Xiaojun Quan. 2021.
\newblock Ubar: Towards fully end-to-end task-oriented dialog system with
  gpt-2.
\newblock In \emph{Proceedings of the AAAI Conference on Artificial
  Intelligence}, volume~35, pages 14230--14238.

\bibitem[{Yu et~al.(2021)Yu, He, Zhang, Du, Pasupat, and
  Li}]{DBLP:conf/naacl/YuHZDPL21}
Dian Yu, Luheng He, Yuan Zhang, Xinya Du, Panupong Pasupat, and Qi~Li. 2021.
\newblock \href {https://doi.org/10.18653/v1/2021.naacl-main.59} {Few-shot
  intent classification and slot filling with retrieved examples}.
\newblock In \emph{Proceedings of the 2021 Conference of the North American
  Chapter of the Association for Computational Linguistics: Human Language
  Technologies, {NAACL-HLT} 2021, Online, June 6-11, 2021}, pages 734--749.
  Association for Computational Linguistics.

\bibitem[{Zang et~al.(2020)Zang, Rastogi, Sunkara, Gupta, Zhang, and
  Chen}]{zang-etal-2020-multiwoz}
Xiaoxue Zang, Abhinav Rastogi, Srinivas Sunkara, Raghav Gupta, Jianguo Zhang,
  and Jindong Chen. 2020.
\newblock \href {https://doi.org/10.18653/v1/2020.nlp4convai-1.13}
  {{M}ulti{WOZ} 2.2 : A dialogue dataset with additional annotation corrections
  and state tracking baselines}.
\newblock In \emph{Proceedings of the 2nd Workshop on Natural Language
  Processing for Conversational AI}, pages 109--117, Online. Association for
  Computational Linguistics.

\bibitem[{Zhang* et~al.(2020)Zhang*, Kishore*, Wu*, Weinberger, and
  Artzi}]{bert-score}
Tianyi Zhang*, Varsha Kishore*, Felix Wu*, Kilian~Q. Weinberger, and Yoav
  Artzi. 2020.
\newblock \href {https://openreview.net/forum?id=SkeHuCVFDr} {Bertscore:
  Evaluating text generation with bert}.
\newblock In \emph{International Conference on Learning Representations}.

\bibitem[{Zhang et~al.(2022)Zhang, Peng, Gao, and Meng}]{zhang2022toward}
Xiaoying Zhang, Baolin Peng, Jianfeng Gao, and Helen Meng. 2022.
\newblock Toward self-learning end-to-end task-oriented dialog systems.
\newblock In \emph{Proceedings of the 23rd Annual Meeting of the Special
  Interest Group on Discourse and Dialogue}, pages 516--530.

\bibitem[{Zhang et~al.(2020)Zhang, Ou, and Yu}]{zhang2020task}
Yichi Zhang, Zhijian Ou, and Zhou Yu. 2020.
\newblock Task-oriented dialog systems that consider multiple appropriate
  responses under the same context.
\newblock In \emph{Proceedings of the AAAI Conference on Artificial
  Intelligence}, volume~34, pages 9604--9611.

\bibitem[{Zhao et~al.(2022)Zhao, Cao, Gupta, Lee, Rastogi, Wang, Soltau,
  Shafran, and Wu}]{zhao2022anytod}
Jeffrey Zhao, Yuan Cao, Raghav Gupta, Harrison Lee, Abhinav Rastogi, Mingqiu
  Wang, Hagen Soltau, Izhak Shafran, and Yonghui Wu. 2022.
\newblock Anytod: A programmable task-oriented dialog system.
\newblock \emph{arXiv preprint arXiv:2212.09939}.

\bibitem[{Zhao and Eskenazi(2018)}]{zhao2018zero}
Tiancheng Zhao and Maxine Eskenazi. 2018.
\newblock Zero-shot dialog generation with cross-domain latent actions.
\newblock \emph{arXiv preprint arXiv:1805.04803}.

\bibitem[{Zhao et~al.(2021)Zhao, Wallace, Feng, Klein, and
  Singh}]{pmlr-v139-zhao21c}
Zihao Zhao, Eric Wallace, Shi Feng, Dan Klein, and Sameer Singh. 2021.
\newblock \href {https://proceedings.mlr.press/v139/zhao21c.html} {Calibrate
  before use: Improving few-shot performance of language models}.
\newblock In \emph{Proceedings of the 38th International Conference on Machine
  Learning}, volume 139 of \emph{Proceedings of Machine Learning Research},
  pages 12697--12706. PMLR.

\end{thebibliography}
\bibliographystyle{acl_natbib}

\newpage
\appendix

\section{Detailed Belief Instructions in DST Prompter}
\label{sec:dst_bi}

\begin{figure}[th]
\centering
\includegraphics[width=1.0\columnwidth]{section/figures/belief_instruction_all_detail.pdf}
\caption{Detailed belief instructions in DST Prompter.}
\label{fig:dst_bi}
\end{figure}

\section{A Formatting Example in Policy Prompter}
\label{sec:format_ex}

\label{sec:format_example}
\begin{figure}[th]
\centering
\includegraphics[width=1.0\columnwidth]{section/figures/policy_prompter_full_ex.pdf}
\caption{A formatting example in Policy Prompter.}
\label{fig:format_ex}
\end{figure}

\section{Implementation Details}
\label{sec:imple_details}
$(\RN{1})$ LLMs: We employ ChatGPT ("gpt-3.5-turbo"), GPT-3.5 ("text-davinci-003") and Codex ("code-davinci-002") as the fixed LLMs to implement the proposed \sptod{}. 
Throughout the evaluation, we set temperature to 0.5.
$(\RN{2})$ DST Prompter -- belief instruction: In the context of multi-domain scenarios, the belief instructions encompassing all domains are incorporated, while solely the target domain's belief instruction is introduced in single-domain settings.
$(\RN{3})$ Policy Prompter -- policy skeleton: For the \texttt{Multiwoz} datasets, we manually construct the policy skeleton through observing a few dialogs in the training corpus, following \citet{mehri2021schema, mosig2020star}. In the case of the \texttt{STAR} corpus, we employ flow chart diagrams and several dialogs to develop the policy skeleton, following the guidelines set forth by \citet{mehri2021schema}. Furthermore, we incorporate the pertinent user template utterance into the system action within the policy skeleton to facilitate the LLM's comprehension of directives, in the absence of belief annotations. The prompt examples for the \texttt{STAR} dataset are shown in Appendix \ref{sec:prompt_example}.

\section{Our Proposed \sptod{} with a Dialog Example}
\label{sec:sptod_ex}

\begin{figure*}[!t]
\centering
\includegraphics[width=0.95\linewidth]{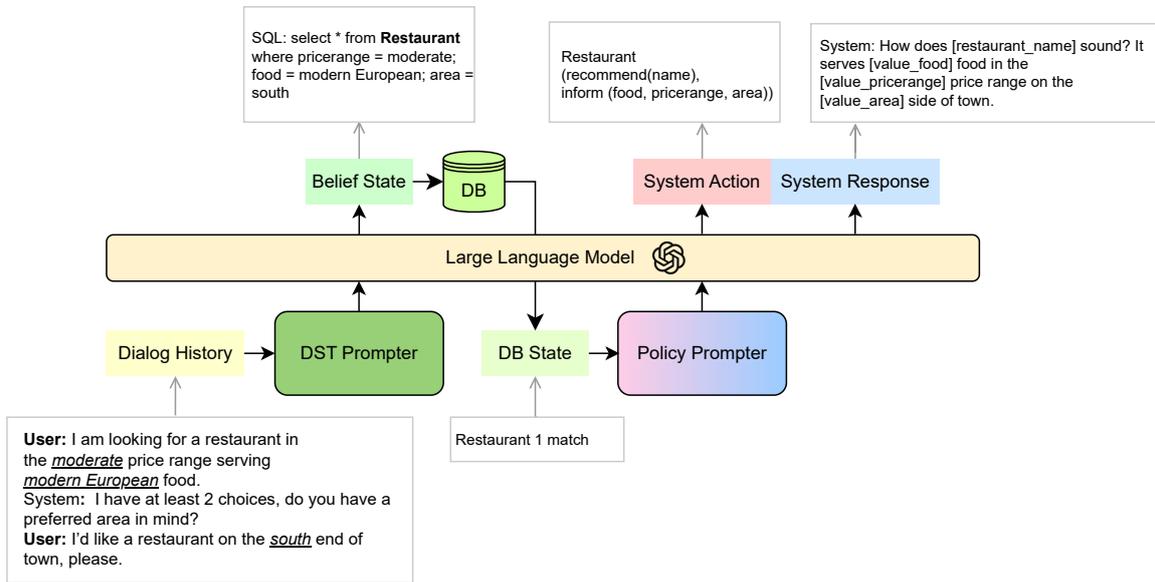}
\caption{Illustration of the proposed \sptod{} with a dialog example. Note that the belief state in the represented in the SQL format, the details of which are described in Section \ref{sec:dst_prop}. }
\label{fig:sptod_ex}
\end{figure*}

\section{Zero-Shot End-to-End Evaluation Results on \texttt{STAR}}
\label{sec:figure_star_acc}

\begin{figure*}[!t]
\centering
\subfigure[Task transfer]{
\begin{minipage}[!t]{0.5\linewidth}
  \centering
  \includegraphics[width=3in]{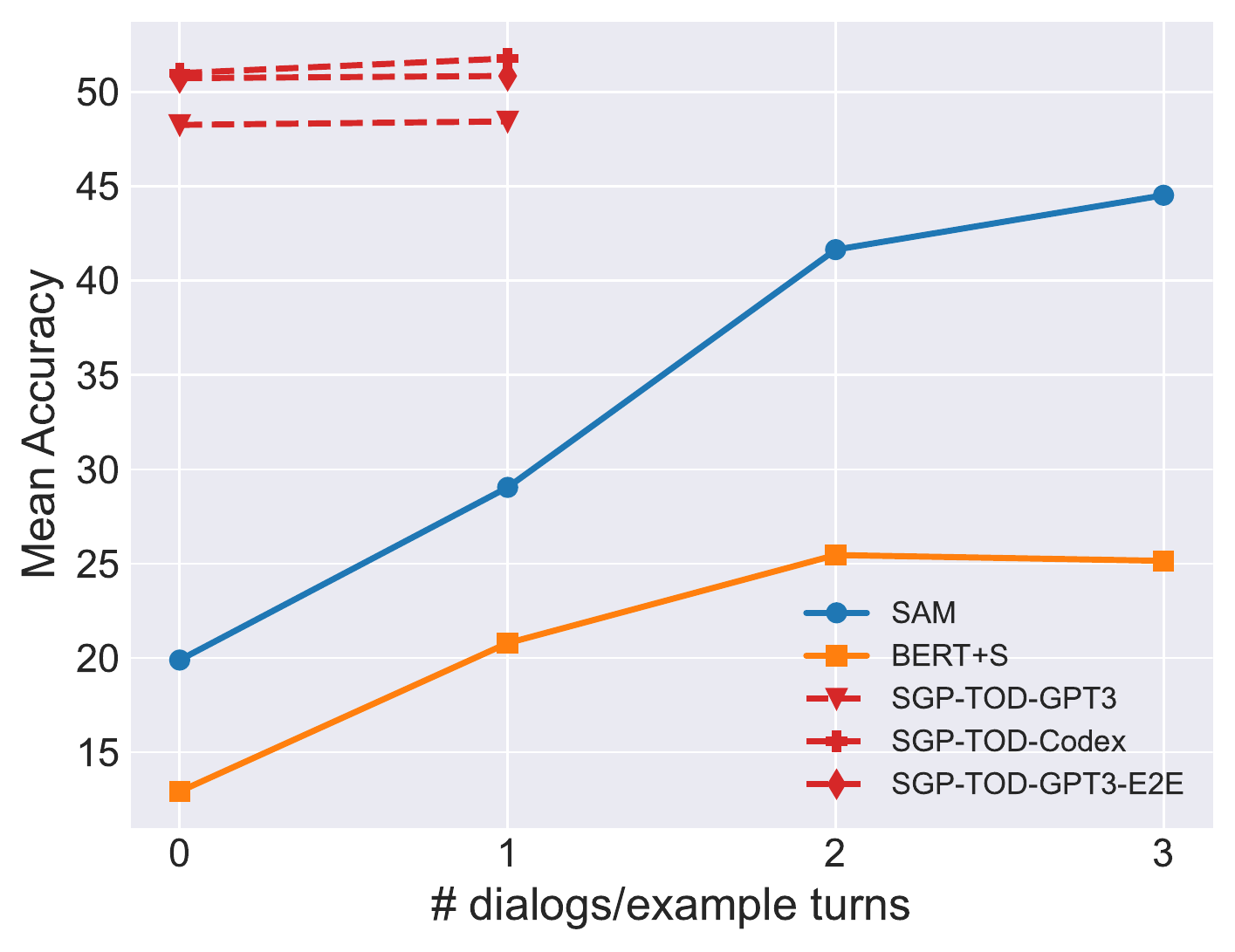}
  \label{fig:cm-bc-s1}
\end{minipage}%
}%
\subfigure[Domain transfer]{
\begin{minipage}[!t]{0.5\linewidth}
  \centering
  \includegraphics[width=3in]{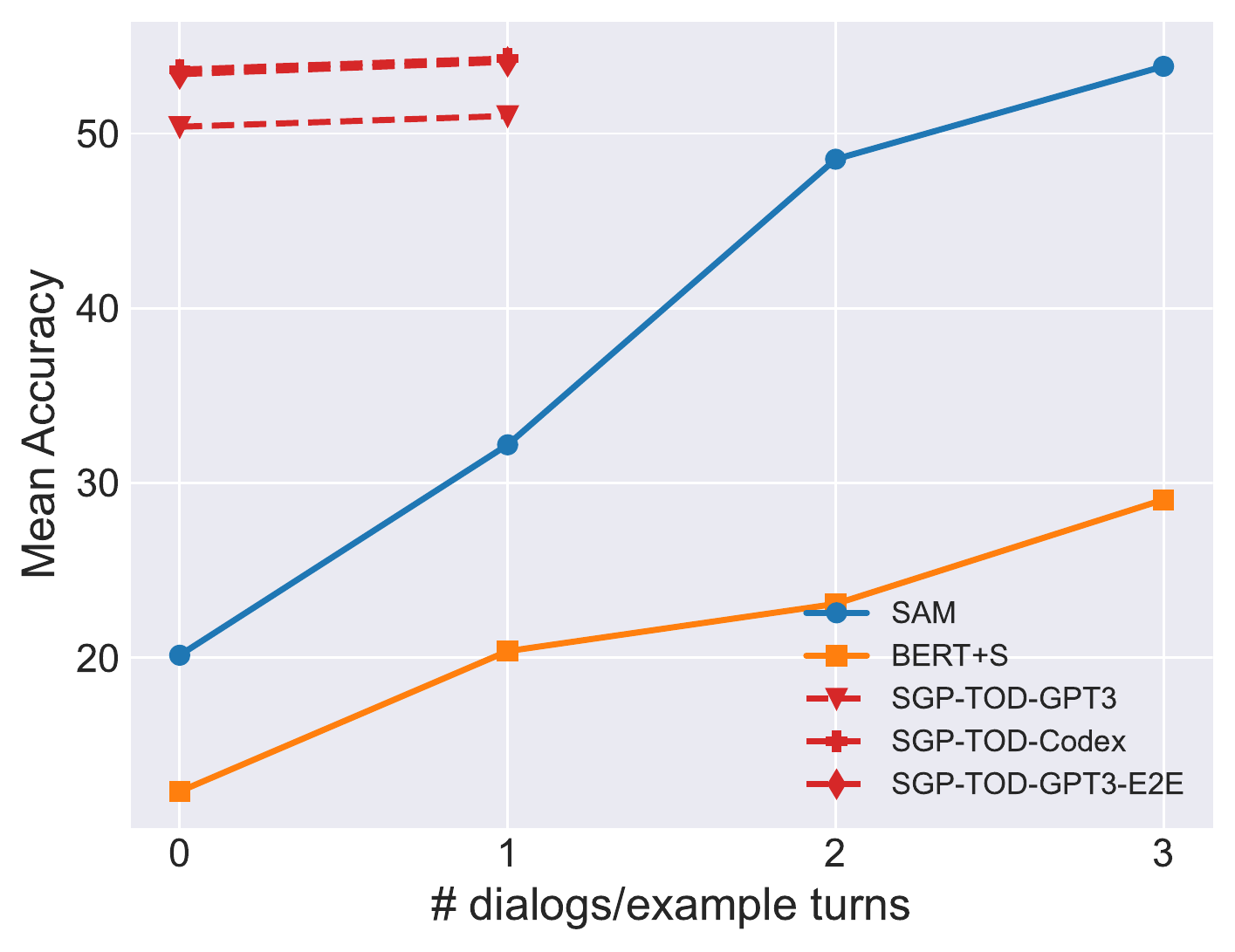}
  \label{fig:cm-bc-s2}
\end{minipage}%
}%
\caption{Zero-shot end-to-end evaluation results on STAR with different number of training dialogs (1, 10, 100, 1,000) / demonstration example turns (1, 10) from source domain/tasks. (Note the numbers are represented in logarithm to base 10.)}
\label{cm-bc}
\end{figure*}

\section{Prompt Examples for \texttt{STAR} Dataset}
\label{sec:prompt_example}

\begin{figure}[!t]
\centering
\includegraphics[width=1.0\columnwidth]{section/figures/ic-codex_star_codex.pdf}
\caption{Policy Prompter of \sptod{} on \texttt{STAR}.}
\label{fig:policy_star}
\end{figure}

\begin{figure}[!t]
\centering
\includegraphics[width=1.0\columnwidth]{section/figures/e2e_policy_prompt.pdf}
\caption{Policy Prompter of \textsc{SGP-TOD-E2E} on \texttt{STAR}.}
\label{fig:e2e_policy_star}
\end{figure}

\section{An Example of Domain Extension}
\label{sec:task_extension}
\begin{figure}[!t]
\centering
\includegraphics[width=0.95\columnwidth]{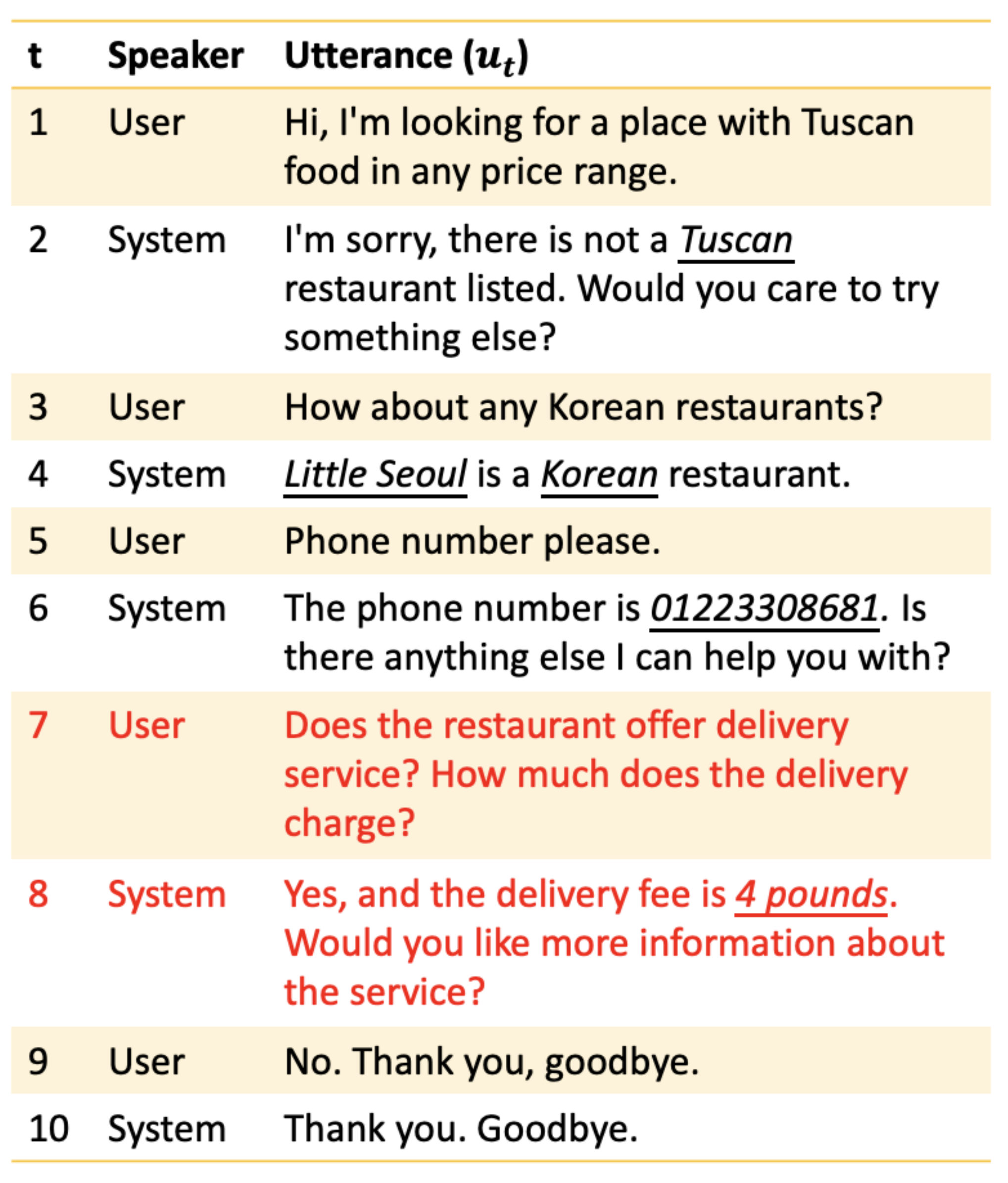}
\caption{A dialog example of domain extension cited from \citet{zhang2022toward}. Task bots need learn to provide pertinent responses concerning the expanded delivery service in additional dialog turns (highlighted in red), as user or business requirements evolve.}
\label{fig:task_extension}
\end{figure}

\begin{figure}[!t]
\centering
\includegraphics[width=0.95\columnwidth]{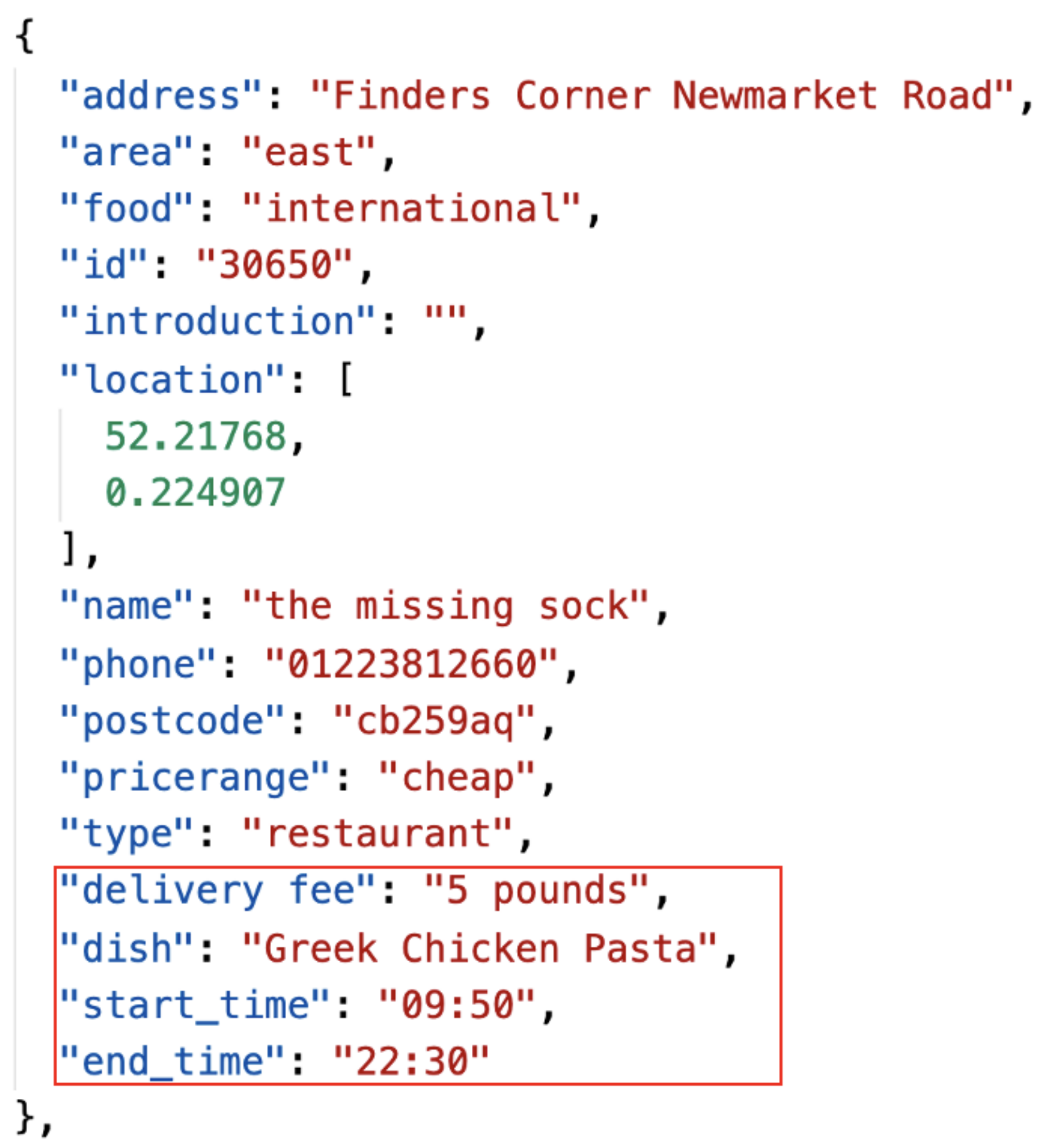}
\caption{An example of \texttt{Restaurant-Ext} DB entry cited from \citet{zhang2022toward}. The supplementary information pertaining to the extended functionality delineated within the red square.}
\label{fig:ex_db}
\end{figure}

\paragraph{A dialog example of domain extension.}
\label{sec:task_extension_dialog}
Figure \ref{fig:task_extension} depicts an example of domain extension.
\paragraph{An example of \texttt{Restaurant-Ext} DB entry.}
An example of \texttt{Restaurant-Ext} DB entry is shown in Figure \ref{fig:ex_db}.
\label{sec:ext_db}


\end{document}